\documentclass{article}

\usepackage[margin=1in]{geometry}
\usepackage[table]{xcolor}% http://ctan.org/pkg/xcolor
\usepackage{graphicx}
\usepackage{booktabs}
\usepackage{amsmath}
\usepackage{amssymb}
\usepackage{amsfonts}
\usepackage{multirow}
\usepackage{verbatim}
\usepackage{caption}
\usepackage{longtable}
\usepackage{supertabular}
\usepackage{float}
\usepackage{lscape}
\usepackage{subfigure}

\usepackage{enumitem}
\usepackage{tablefootnote}
\usepackage[round,semicolon]{natbib}
\usepackage{xcolor}
\usepackage{xspace}
\usepackage{textcomp}
\usepackage{makecell}
\usepackage{multirow}
\usepackage{lscape} 
\usepackage{siunitx}

\usepackage[]{mdframed}

\usepackage[subfigure]{tocloft}

\usepackage{amssymb}% http://ctan.org/pkg/amssymb
\usepackage{pifont}% http://ctan.org/pkg/pifont
\usepackage{scrextend}

\newcommand*\colourcheck[1]{%
  \expandafter\newcommand\csname #1check\endcsname{\textcolor{#1}{\ding{52}}}%
}
\newcommand*\colourcross[1]{%
  \expandafter\newcommand\csname #1cross\endcsname{\textcolor{#1}{\ding{55}}}%
}
\colourcheck{blue}
\colourcheck{red}
\colourcross{red}

\newcommand\greencheck{\textcolor{checkmarkcolor}{\ding{52}}}

\usepackage{array}
\usepackage{tgpagella}
\usepackage{latexsym}
\usepackage[T1]{fontenc}
\usepackage[utf8]{inputenc}
\usepackage{microtype}
\definecolor{mydarkblue}{rgb}{0,0.08,0.45}
\PassOptionsToPackage{hyphens}{url}\usepackage[colorlinks,citecolor=mydarkblue,urlcolor=mydarkblue,linkcolor=mydarkblue,linktocpage=true]{hyperref}
\usepackage{url}            % simple URL typesetting
\usepackage{nicefrac}       % compact symbols for 1/2, etc.
\usepackage{changepage}
\usepackage{xargs}          % Use more than one optional parameter in a new commands
\usepackage{wrapfig,lipsum,booktabs}
\usepackage{longtable}
\usepackage{endnotes}

\usepackage{pgfplots}
\usetikzlibrary{pgfplots.groupplots}
\pgfplotsset{compat=1.3}
\usepackage{tikz}
\usetikzlibrary{patterns}

\usepackage[most]{tcolorbox}

\usepackage[capitalize,noabbrev]{cleveref}
\crefname{section}{Section}{\S\S}
\Crefname{section}{Section}{\S\S}
\crefname{table}{Table}{Tables}
\crefname{figure}{Figure}{Figures}
\crefname{algorithm}{Algorithm}{}
\crefname{equation}{eq.}{}
\crefname{appendix}{Appendix}{}
\crefformat{section}{Section #2#1#3}
\usepackage{multicol}

\definecolor{battleshipgrey}{rgb}{0.3, 0.3, 0.3}
\definecolor{brilliantrose}{rgb}{1.0, 0.33, 0.64}
\definecolor{americanrose}{rgb}{1.0, 0.01, 0.24}
\definecolor{jweigreen}{rgb}{0,0.45,0.24}
\definecolor{bluegray}{rgb}{0.1, 0.1, 0.4}
\definecolor{ao(english)}{rgb}{0.0, 0.5, 0.0}
\definecolor{blanchedalmond}{rgb}{1.0, 0.92, 0.8}
\definecolor{atomictangerine}{rgb}{1.0, 0.6, 0.4}
\definecolor{chocolate(web)}{rgb}{0.82, 0.41, 0.12}
\definecolor{bananayellow}{rgb}{1.0, 0.88, 0.21}
\definecolor{goldenbrown}{rgb}{0.6, 0.4, 0.08}
\definecolor{aliceblue}{rgb}{0.94, 0.97, 1.0}
\definecolor{beige}{rgb}{0.96, 0.96, 0.86}
\definecolor{babyblue}{rgb}{0.54, 0.81, 0.94}
\definecolor{camel}{rgb}{0.76, 0.6, 0.42}
\definecolor{cinnamon}{rgb}{0.82, 0.41, 0.12}
\definecolor{deepskyblue}{rgb}{0.0, 0.75, 1.0}
\definecolor{frenchblue}{rgb}{0.0, 0.45, 0.73}
\definecolor{classicrose}{rgb}{0.98, 0.8, 0.91}
\definecolor{frenchrose}{rgb}{0.96, 0.29, 0.54}
\definecolor{frenchlilac}{rgb}{0.53, 0.38, 0.56}
\definecolor{frenchbeige}{rgb}{0.65, 0.48, 0.36}
\definecolor{checkmarkcolor}{RGB}{130, 212, 83}

\definecolor{forestgreen}{HTML}{2e7d43}
\definecolor{goldenrod}{HTML}{daa520}
\definecolor{color1}{HTML}{FF9999}
\definecolor{color2}{HTML}{FF6666}
\definecolor{color3}{HTML}{FF3333}
\definecolor{color4}{HTML}{E60000}
\definecolor{color5}{HTML}{B30000}
\definecolor{color6}{HTML}{8CD98C}
\definecolor{color7}{HTML}{53c653}
\definecolor{color8}{HTML}{39ac39}
\definecolor{color9}{HTML}{2d862d}
\definecolor{color10}{HTML}{206020}
\definecolor{color11}{HTML}{cca300}

\newcommand{\smalLM}[0]{\textsc{LM-Small}}
\newcommand{\bigLM}[0]{\textsc{LM-XL}}

\usepackage{minitoc}

\usepackage[bottom,symbol]{footmisc}

\title{
\vspace{-10mm}
\textbf{
A Pretrainer's Guide to Training Data: \\Measuring the Effects of Data Age, Domain Coverage, Quality, \& Toxicity
}
\vspace{-3mm}
  }

\renewcommand\footnotemark{}
\newcommand\tinyspace{\hspace{0.12em}}
\author{
\normalsize{}
\textbf{Shayne Longpre\tinyspace$^{\mathbf{1}\tinyspace\diamondsuit\text{\tinyspace*}}$\thanks{*\,Work completed while a Student Researcher at Google Research. \newline \hspace*{1.5em} \textsuperscript{\textdagger}\,Work completed at Google Research.\newline \hspace*{1.5em} $^\diamondsuit$ \,Core contributor. Correspondence: \url{slongpre@media.mit.edu}}}\hspace{5mm}
\textbf{Gregory Yauney\tinyspace$^{\mathbf{2}\tinyspace\diamondsuit\text{\tinyspace*}}$} \hspace{5mm} 
\textbf{Emily Reif\tinyspace$^{\mathbf{3}\tinyspace\diamondsuit}$} \hspace{5mm}
\textbf{Katherine Lee\tinyspace$^{\mathbf{2,3}\tinyspace\diamondsuit}$} \hspace{5mm}
\\
\normalsize{}
\textbf{Adam Roberts\tinyspace$^{\mathbf{3}}$} \hspace{5mm} 
\textbf{Barret Zoph\tinyspace$^{\mathbf{4}\tinyspace\text{\textdagger}}$} \hspace{5mm}
\textbf{Denny Zhou\tinyspace$^{\mathbf{3}}$} \hspace{5mm} 
\textbf{Jason Wei\tinyspace$^{\mathbf{4}\tinyspace\text{\textdagger}}$} \hspace{5mm}
\textbf{Kevin Robinson\tinyspace$^{\mathbf{3}}$} \hspace{5mm}
\\
\normalsize{}
\textbf{David Mimno\tinyspace$^{\mathbf{2}\tinyspace\diamondsuit}$} \hspace{5mm} 
\textbf{Daphne Ippolito\tinyspace$^{\mathbf{3}}$\tinyspace$^{\diamondsuit}$} \hspace{4mm}
\\
\\
\normalsize{}
$^{1}$ MIT
\hspace{5mm}
$^{2}$ Cornell University
\hspace{5mm}
$^{3}$ Google Research
\hspace{5mm}
$^{4}$ OpenAI
\vspace{-4mm}
}

\date{}

\begin{document}

%\doparttoc % Tell to minitoc to generate a toc for the parts
%\faketableofcontents % Run a fake tableofcontents command for the partocs
% \part{} % Start the document part
% \parttoc{} % Insert the document TOC

\maketitle

\begin{abstract}
\noindent 
Pretraining is the preliminary and fundamental step in developing capable language models (LM).
Despite this, pretraining data design is critically under-documented and often guided by empirically unsupported intuitions.
To address this, we pretrain 28 1.5B parameter decoder-only models, training on data curated (1) at different times, (2) with varying toxicity and quality filters, and (3) with different domain compositions.
% \noindent
First, we quantify the effect of pretraining data age.
A temporal shift between evaluation data and pretraining data leads to performance degradation, which is not overcome by finetuning.
% This phenomenon is exacerbated by model scale.
Second, we explore the effect of quality and toxicity filters, showing a trade-off between performance on standard benchmarks and risk of toxic generations.
Our findings indicate there does not exist a one-size-fits-all solution to filtering training data.
We also find that the effects of different types of filtering are not predictable from text domain characteristics.
% Together these results indicate one size filter does not fit all objectives, and future work should explore more nuanced definitions of quality and relevance.
% Second, we recommend quality filters or inverse toxicity filters, which provide a trade-off: broad performance boosts, but at the expense of toxic generation.
% Our analysis also shows the effects of different types of filtering were not predictable from text domain characteristics.
Lastly, we empirically validate that the inclusion of heterogeneous data sources, like books and web, is broadly beneficial and warrants greater prioritization.
% , though omitting any text domains is detrimental.
% dei: Detrimental to what? Since we don't want the abstract to be a page+ long, i just commented this out.
These findings constitute the largest set of experiments to validate, quantify, and expose many undocumented intuitions about text pretraining, which we hope will help support more informed data-centric decisions in LM development.
\end{abstract}

\vspace{1em}

\begin{figure}[ht]
    \centering
    \includegraphics[width=0.99 \linewidth]{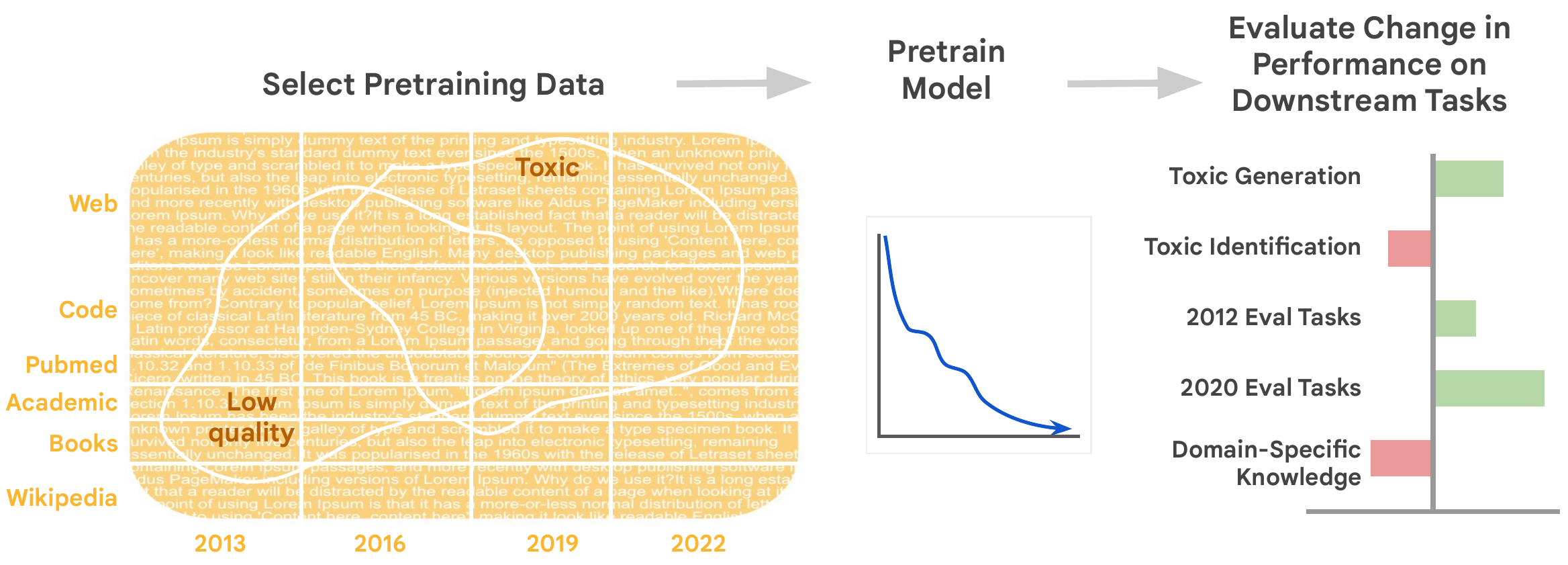}
    \caption{
    \small \textbf{The experimental pretraining curation pipeline} includes three steps: sub-selecting data from C4 or the Pile, pretraining a language model, and evaluating its change in performance over several benchmarks.}
    \vspace{-3mm}
    \label{fig:intro}
\end{figure}
\thispagestyle{empty}
\newpage
%\tableofcontents

\doparttoc % Tell to minitoc to generate a toc for the parts
\faketableofcontents % Run a fake tableofcontents command for the partocs
\noptcrule 
\tightmtctrue
\renewcommand{\ptctitle}{Contents}
\part{} % Start the document part
\parttoc{} % Insert the document TOC
\thispagestyle{empty}

\newpage

% \clearpage
\vspace{-1mm}
\section{Introduction}
\vspace{-3mm}

The strong performance \citep{chowdhery2022palm, chinchilla2022implications,openai2023gpt4,google2023}, and emergent abilities \citep{wei2022emergent} of modern language models (LMs) depend on self-supervised pretraining on massive text datasets.
All model developers implicitly or explicitly decide the composition of these datasets: what data sources to include, whether to filter for attributes such as quality and toxicity, and when to gather new documents.
While many of the most prominent models do not document their curation procedures \citep{openai2023gpt4,google2023}, or only document \textit{which} procedures they used \citep{brown2020language,chinchilla2022implications,scao2022bloom,touvron2023llama}, they rarely document \textit{why} they chose those protocols or what effect they had.
This documentation debt leaves practitioners to be guided by intuitions and precedents, neither thoroughly evaluated \citep{bandy2021addressing, sambasivan2021everyone}.
Given the outsized and fundamental role of pretraining data in modern LMs, we believe this neglectful practice has detracted from responsible data use and hampered effective model development \citep{rogers2021changing, gebru2021datasheets, bender2018data}.
% Our goal is to provide model builders with insights to effectively develop corpora for their target applications, and to help downstream users to choose amongst model candidates for their application.
% In this work, we systematically test how dataset decisions affect model performance.

Among the small number of general-purpose LMs dominating community use and discussion, the prevailing focus has been on the scale of pretraining data and number of optimization steps \citep{brown2020language,chinchilla2022implications,google2023}.
In this work, we systematically test how common data design decisions affect model performance---specifically: the time of collection, content filtering strategy (toxicity/quality), and domain composition.
We study the impacts in two ways. 
First, we present observational measurements of the effect of existing quality and toxicity filtering methods (\cref{sec:data-analysis}). 
We document how these filters affect a range of characteristics in two major pretraining datasets, C4 \citep{raffel2020exploring} and the Pile \citep{gao2020pile}.
Second, we rigorously evaluate these dataset design decisions on downstream tasks.
This is done by evaluating decoder-only autoregressive LMs each pretrained on a dataset modified along one dimension of time, toxicity, quality, or domain composition. 
Our contributions are summarized as findings and recommendations to model developers.

\textbf{The Age of a Dataset (\cref{sec:time}).} We see performance degradation if evaluation data is \emph{either} before or \emph{after} pretraining data collection, and this deficit isn't overcome with substantial finetuning.
Further, this phenomenon is exacerbated in larger models. 
While rarely acknowledged, we show its effect can meaningfully complicate comparisons between new and old models, depending on the age of the evaluation dataset.
% , as evaluation datasets also depend on their collection time.

\textbf{Quality and Toxicity Filters (\cref{sec:quality-toxicity}).}  Filtering for document quality and toxicity have significant but opposite effects on model behaviour. Quality filtering, removing low-quality text, substantially increases both toxic generation and downstream performance across tasks we tested, \emph{despite reducing the amount of training data}.
On the other hand, removing toxic data trades-off fewer toxic generations for reduced generalization performance.
Inverse toxicity filters, which remove the least toxic content, demonstrate targeted benefits. 
% actually \textbf{improve performance on toxicity identification}.
Lastly, evaluation on datasets with high quality text aren't necessarily improved by removing low-quality text from the dataset. 
% Lastly, the content classified as highest quality isn't improved the most by quality filtering, nor is the lowest quality improved the least.
Performance effects due to quality filtering are mostly positive, but the benefits are not predictable from text characteristics. 
These findings demonstrate that \emph{one size (filter) does not fit all}, and there is a need for practitioners to develop more targeted quality or inverse toxicity filters for their tasks.

\textbf{Domain Compositions (\cref{sec:domain-comp}).} The best performing domains comprise high-quality (Books) and heterogeneous (Web) data, corroborating \citet{brown2020language, chowdhery2022palm, xie2023doremi}.
However, these text sources contribute most to toxic generation.
Still, we found that the benefits of training on these data sources is often greater than data collection for a targeted domain, and so recommend practitioners focus future collection on more books and diverse web data.
Additionally, our best performing models still use all data sources (even at the relatively small scale of 1.5B parameters); thus, we recommend practitioners generously include data sources less relevant to their downstream tasks \citep{madaan2022language}.

To our knowledge, these experiments constitute the largest publicly documented LM data curation study, spanning 28 1.5B parameter models.
Their findings \emph{empirically quantify, validate, and, occasionally, challenge} the entrenched set of under-examined pretraining assumptions; which we believe justifies their computational cost (\cref{sec:limitations}).
As the majority of the community has adopted a small set of models for most research and applications (BERT, T5, GPT-2, GPT-3), pretraining data curation decision have long-term ramifications.
We hope these results better inform model developers training the next wave of LMs.

\vspace{-3mm}
\section{Methodology}

We measure how pretraining data curation choices affect downstream performance. \cref{fig:intro} illustrates our approach: each experiment starts with a pretraining dataset, applies a filter that removes documents, pretrains a language model on the curated dataset, and finally evaluates the model on downstream tasks.

\begin{table}[t]
    \centering
    \small
\caption{A list of \textbf{well-known language models and a quantitative breakdown of their pretraining data}, including represented domains; if the Pile or C4 are used, the percent of multilingual (M-L) data (meaning non-English and non-code); if Toxicity or Quality data filters were used, as either automatic Heuristics (H) or Classifiers (C); if the dataset is public (Pub), and what year the data was collected up to.
If a dataset is ``Part'' public, then all of its constituent corpora are public, but not the final mixture.
In Represented Domains, extended from \citep{zhao2023survey}, Web includes the Common Crawl and other web scrapes; Dialog includes forum, social media and conversations; Academic includes research papers, textbooks, and mathematics.
}
\label{tab:model_data_survey}
% \begin{tabular}{p{0.5in}|llllllp{0.8in}l|p{2in}ll|l}
\begin{tabular}{l|rrrrrc|ccc|cc|cc}
\toprule
&\multicolumn{6}{c|}{\textsc{Represented Domains (\%)}}& & & &\multicolumn{2}{c|}{\textsc{Filters}}&\multicolumn{2}{c}{\textsc{Data}}\\
\textsc{Model} & \textsc{Wiki} & \textsc{Web} & \textsc{Books} & \textsc{Dialog} & \textsc{Code} & \textsc{Acad} & \textsc{Pile} & \textsc{C4} & \textsc{M-L} & \textsc{Tox} & \textsc{Qual} & \textsc{Pub} & \textsc{Year}\\
\midrule
\textsc{Bert} & \cellcolor{forestgreen!76}76 & & \cellcolor{forestgreen!24}24 & &  &  & \redcross & \redcross & & & \textsc{H} & Part & 2018\\
\textsc{GPT-2} &  & \cellcolor{forestgreen!100}100 &  &  &  & & \redcross & \redcross &  & & \textsc{H} & Part & 2019\\
\textsc{RoBerta} & \cellcolor{forestgreen!7}7 & \cellcolor{forestgreen!90}90 & \cellcolor{forestgreen!3}3 &  &  &  & \redcross & \greencheck & & & \textsc{H} & Part & 2019\\
\textsc{XLNet} & \cellcolor{forestgreen!8}8 & \cellcolor{forestgreen!89}89 & \cellcolor{forestgreen!3}3 &  &  &  & \redcross & \greencheck &  & & \textsc{H} & Part & 2019\\
\textsc{T5} & \cellcolor{forestgreen!1}<1 & \cellcolor{forestgreen!99}99 &  &  &  & & \redcross & \greencheck &  & \textsc{H} & \textsc{H} & \greencheck & 2019 \\
% July 2020 release, 7\% multilingual text according to section 3.3
\textsc{GPT-3} & \cellcolor{forestgreen!3}3 & \cellcolor{forestgreen!82}82 & \cellcolor{forestgreen!16}16 & & & & \redcross & \greencheck & \cellcolor{goldenrod!7}7\% &  & \textsc{C} & \redcross & 2021 \\
% GPT-J: Aug 2021. Pile: Dec 2020.
\textsc{GPT-J/Neo} & \cellcolor{forestgreen!1.5}1.5 & \cellcolor{forestgreen!38}38 & \cellcolor{forestgreen!15}15 & \cellcolor{forestgreen!5}4.5 & \cellcolor{forestgreen!13}13 & \cellcolor{forestgreen!28}28 & \greencheck & Part & & & \textsc{C} & \greencheck & 2020 \\
% Dec 2021. Tox filter because they cite T5's C4 directly.
\textsc{GLaM} & \cellcolor{forestgreen!6}6 & \cellcolor{forestgreen!46}46 & \cellcolor{forestgreen!20}20 & \cellcolor{forestgreen!28}28 &  & & \redcross & \greencheck &  & & \textsc{C} & \redcross & 2021\\
% Jan 20, 2022.
\textsc{LaMDA} & \cellcolor{forestgreen!13}13 & \cellcolor{forestgreen!24}24 &  & \cellcolor{forestgreen!50}50 & \cellcolor{forestgreen!13}13 & & \greencheck & \greencheck & \cellcolor{goldenrod!10}10\% & \textsc{C} & \textsc{C} & \redcross & 2021\\
% Feb 2022 Release. 
% Filtering: (section B.2) and "we
% filtered out all files larger than 1MB or with lines longer than 1000 characters, to exclude automatically
% generated code. We also removed duplicates of the same file, ignoring whitespace in comparisons"
\textsc{AlphaCode} &  &  &  & & \cellcolor{forestgreen!100}100 &  & \redcross & \redcross &  & & \textsc{H} & \redcross & 2021 \\
% March 2022
\textsc{CodeGen} & \cellcolor{forestgreen!1}1 & \cellcolor{forestgreen!24}24 & \cellcolor{forestgreen!10}10 & \cellcolor{forestgreen!3}3 & \cellcolor{forestgreen!40}40 & \cellcolor{forestgreen!22}22 & \greencheck & Part & & & \textsc{H} & Part & 2020 \\
% march 2022 release
\textsc{Chinchilla} & \cellcolor{forestgreen!1}1 & \cellcolor{forestgreen!65}65 & \cellcolor{forestgreen!10}10 & & \cellcolor{forestgreen!4}4 & & \greencheck & \greencheck &  & \textsc{H} & \textsc{C} & \redcross & 2021 \\
% June 2022. Filters: Uses 5% PaLM data (same filters), and 95% acad.
\textsc{Minerva} & \cellcolor{forestgreen!1}<1 & \cellcolor{forestgreen!2}1.5 & \cellcolor{forestgreen!1}<1 & \cellcolor{forestgreen!2.5}2.5 & \cellcolor{forestgreen!1}<1 & \cellcolor{forestgreen!95}95 & \greencheck & \greencheck & \cellcolor{goldenrod!1}<1\% & & \textsc{C} & \redcross & 2022 \\
% Published Nov 2022. Released July 2022. 29% English data.
% 62% of docs come from ROOTS sec 2: (11% code, 10% academic, 10% books, 5% wiki) , then 38% OSCAR (ML CC)
\textsc{BLOOM} & \cellcolor{forestgreen!5}5 & \cellcolor{forestgreen!60}60 & \cellcolor{forestgreen!10}10 & \cellcolor{forestgreen!5}5 & \cellcolor{forestgreen!10}10 & \cellcolor{forestgreen!10}10 & \greencheck & \greencheck & \cellcolor{goldenrod!71}71\% & \textsc{H} & \textsc{C} & Part & 2021 \\
% October 2022 release, Appendix D says data collected up to 2021. 22% multilingual
\textsc{PaLM} & \cellcolor{forestgreen!4}4 & \cellcolor{forestgreen!28}28 & \cellcolor{forestgreen!13}13 & \cellcolor{forestgreen!50}50 & \cellcolor{forestgreen!5}5 & & \redcross & \greencheck & \cellcolor{goldenrod!22}22\% &  & \textsc{C} & \redcross & 2021 \\
% Nov 16 2022. Filters: We apply several quality filters, including excluding papers from journals with certain keywords, and also
% excluding papers with a low journal impact factor.
\textsc{Galactica} & \cellcolor{forestgreen!1}1 & \cellcolor{forestgreen!7}7 & \cellcolor{forestgreen!1}1 & & \cellcolor{forestgreen!7}7 & \cellcolor{forestgreen!84}84 & \greencheck & Part & & & \textsc{H} & Part & 2022 \\
% May 2022, appendix C.2 doesn't give enough info for percentages
% \textsc{OPT} &  &  &  &  &  & & Pile & \ &  &  &  & ? \\
% Feb 2023 release. data up to 2020. Used multilingual wikipedia (~4% all data)
\textsc{LLAMA} & \cellcolor{forestgreen!5}4.5 & \cellcolor{forestgreen!82}82 & \cellcolor{forestgreen!5}4.5 & \cellcolor{forestgreen!2}2 & \cellcolor{forestgreen!5}4.5 & \cellcolor{forestgreen!3}2.5 & Part & \greencheck & \cellcolor{goldenrod!4}4\% & & \textsc{C} & Part & 2020 \\
\bottomrule
\end{tabular}
\end{table}

\vspace{-3mm}
\subsection{Pretraining Datasets}

We begin with two common, publicly available pretraining datasets: C4 \citep{raffel2020exploring} and the Pile \citep{gao2020pile}.
Both have received basic initial heuristic filtering for English language and content quality.
We further deduplicate both datasets using the approximate deduplication method described in \citet{lee2022deduplicating}.
% Additional details on pretraining are specified in \cref{app:pretrain-hps}.

\vspace{-3mm}
\paragraph{C4 \citep{raffel2020exploring}} The English Colossal Clean Crawled Corpus (C4) is a snapshot of Common Crawl from 2019, which includes a mix of news, legal, wikipedia, and generic web documents \citep{dodge2021documenting}, filtered for well-formed English text.\footnote{\url{https://commoncrawl.org/}}
While the original version of C4 filtered out any documents containing words from a ``bad words list'', our version does not.
C4 remains one of the most widely adopted fully open source datasets for textual training, given its permissive license.
It is a key component of many LMs, as shown in \cref{tab:model_data_survey}.
% pretraining in XLNet \citep{yang2020xlnet}, RoBERTa \citep{liu2019roberta}, T5 \citep{raffel2020exploring}, STMoE \citep{zoph2022stmoe}, UL2 \citep{tay2023ul2}, and the GPT-3 model series \citep{brown2020language}.

\vspace{-3mm}
\paragraph{The Pile \citep{gao2020pile}} is an 800GB dataset consisting of data from 22 sources.
These include a Common Crawl web scrape as well as more diverse collections of academic, books, coding, medical, legal and social sources (see \cref{tab:pile-partitions}), which more closely resemble the reported data sources in larger non-open source models like PaLM \citep{chowdhery2022palm}, Chinchilla \citep{hoffmann2022training}, and the GPT-3 series \citep{brown2020language}.
Note that the Pile's corpora composition was manually selected, and some options were excluded on the grounds of being too toxic or explicit.
% The Pile is a primary pretraining data source in OPT \citep{zhang2022opt}, GPT-J \citep{gpt-j}, and GPT-Neo \citep{gpt-neo}.

\vspace{-3mm}
\subsection{Data Curation Choices}
\label{sec:data-filters}

We evaluate variations in the pretraining data based on three categories of interventions.

\vspace{-3mm}
\paragraph{Dataset Age}
We create new versions of C4 by regenerating snapshots of the Common Crawl from different years (see \cref{fig:dates-histogram}).
Multiple time-based collections are not available for the Pile. 

\vspace{-3mm}
\paragraph{Domain Filtering}
Both C4 and the Pile draw from multiple distinct data sources, but the Pile explicitly delineates 22 distinct sources from web pages, wikipedia articles, code repositories, online forums, legal texts, and research paper archives.
To control for the topical content of the pretraining collection, we selectively remove documents from different domains (see \cref{tab:pile-partitions}).
% C4 and the Pile do not contain precisely the same data sources, but we are able to make an approximate matching between sources.

\vspace{-3mm}
\paragraph{Content Filtering}
Datasets derived from the Common Crawl and other weakly curated internet sources tend to contain large amounts of low-quality, toxic, or offensive content.
As a result, curators often apply content-based filters.
Deciding what to include and what not to include is a challenging and context-dependent problem: A ``high-quality'' Reddit post does not look like a ``high-quality`` academic paper; and even with academic papers, quality measured by peer review has high variance \citep{cortes2021inconsistency}.

There are several approaches to determining document appropriateness.
The simplest filters use features such as sentence length, presence of stopwords and punctuation, and repetitiousness to identify pages that do not contain usable text \citep{rae2021scaling,yang2019xlnet,laurenccon2022bigscience,zhang2022opt}.
Negatively-defined filters identify a category of text to be removed, and assume that everything else is usable.
For example, \citet{raffel2020exploring} remove documents that contain words from a list of ``bad words''.
Positively-defined filters identify a category of text to keep, and %assume that everything else should be removed 
remove everything else \citep{du_glam_2021,touvron2023llama,brown2020language}.

In this work, we evaluate the impact of two document-level, classifier-based  filters that have been used widely in the development of state-of-the-art language models.
These include negatively-defined, \textit{toxic} content (text that is profane, explicit, insulting, or threatening) and postively-defined \textit{quality} content (text similar to known ``high-quality'' sources).
It is important to emphasize that we do not have ground truth: for the purposes of this paper we will use the description \textit{toxic} or \textit{quality} to refer to a document that triggers one of these automated classifiers, \textit{not} to indicate a document that achieves those characteristics for a human reader.

\vspace{-3mm}
\paragraph{Quality Filters}
% Most recent language models use positively-defined quality filters, usually to crawled web pages.
Most recent language models create quality classifiers to distinguish between %reference 
``high-quality'' corpora and other documents (\cref{tab:model_data_survey}).
These are usually then applied to crawled web pages.
Examples of high-quality reference corpora are (1) Wikipedia, WebText and books for GPT-3 \citep{brown2020language}, (2) Wikipedia, books and a few selected websites for PaLM \citep{chowdhery2022palm} and GLaM \citep{du_glam_2021},
and (3) pages used as references in Wikipedia for LLaMA \citep{touvron2023llama}.
In our work, we use the classifier employed by PaLM and GLaM, which assigns each document a score from 0 (high quality) to 1 (low quality).
We experiment with removing documents that fall above four quality thresholds: $0.975$, $0.95$, $0.9$, $0.7$, along with an inverse filter that instead removes the \textit{highest} quality documents \emph{below} a threshold.

\vspace{-3mm}
\paragraph{Toxicity Filters}
To identify toxic content, we use Jigsaw's Perspective API \footnote{\url{https://www.perspectiveapi.com}}, which was trained on comments from online forums and assigns toxicity scores based on whether annotators found the comment to contain profanity/obscenity, identity-based negativity, insults, or threats.
While the Perspective API, as with any classifier, has been shown to be imperfect---it falsely labels some neutral text as toxic and its training data reflects the normative values of its annotators---it has been shown to be far more accurate than heuristic and rule-based classifiers \citep{friedl2023dis,gargee2022analyzing,lees2022perspective}.

The Perspective API outputs a score from 0 (unlikely to be toxic) to 1 (very likely to be toxic).
The documentation recommends using a score threshold of anywhere from 0.3 to 0.9 to filter documents, depending on the practitioner's goals.\footnote{See \url{https://developers.perspectiveapi.com/s/about-the-api-score}}
We experiment with removing documents with scores above five different toxicity threshold values $0.95$, $0.9$, $0.7$, $0.5$, and $0.3$. Documents above a given threshold are filtered out, along with an inverse filter that removes documents with the \textit{least} predicted toxicity \textit{below} a threshold.

In addition to the classifier-based filter, we also experiment with the $n$-gram based filter used by \citet{raffel2020exploring} in the original version of the C4 dataset.
This filter removes all documents that contain any word present in the ``List of Dirty, Naughty, Obscene, or Otherwise Bad Words''.\footnote{\url{https://github.com/LDNOOBW/List-of-Dirty-Naughty-Obscene-and-Otherwise-Bad-Words}}

% \subsection{Finetuning \& Evaluation}
\vspace{-3mm}
\subsection{Evaluation}
\label{sec:finetune-eval}

To measure the effects of time, topic and toxicity, we evaluate pretrained models on English-language tasks for toxicity identification, toxic generation, dozens of question-answering (QA) tasks from diverse domains, and several tasks with temporal annotations.
In choosing evaluations, we compare the general utility of the different models, as well as their performance on tasks we expect to be influenced by the dataset characteristics being ablated.
Since we are comparing the performance of different pretrained models, we evaluate the performance of each pretrained model on downstream tasks by finetuning the model on the relevant dataset for each task
% Although we are evaluating finetuned models, for each task each pretrained model is finetuned with the same training data 
and evaluated on the same testing data (using the default splits for each task unless otherwise noted).
As a result, any \emph{systematic} differences between finetuned results can only be attributable to differences in pretraining.
%Unless otherwise noted, for all benchmark tasks we finetune the pretrained model on the task's train set and report performance on its test set.
%It is important to note that the reported results directly measure the only independent variable: the effect of the pretraining data.
% It is important to note that the reported results are thus a measure of the impact of pretraining even in the presence of additional context-specific finetuning.
For all tasks we report mean performance relative to a baseline, usually the performance of models trained on an unfiltered dataset.

\vspace{-3mm}
\paragraph{Evaluating Domain Generalization}
% As is common practice, we evaluate the general utility of a language model by its performance on the SuperGLUE natural language understanding benchmark \citep{wang2019superglue}.
% As is standard practice, we evalaute NLU using the SuperGLUE benchmark \citep{wang2019superglue}.
We evaluate on the union of two question-answering benchmarks: Machine Reading for Question Answering (MRQA) \citep{fisch2019mrqa} and UnifiedQA \citep{khashabi-etal-2020-unifiedqa}, which together consist of 30 unique QA datasets.
These QA datasets span a range of domains, allowing us to measure the impact of topic alignment (see \cref{tab:qa-partitions}).

\vspace{-3mm}
\paragraph{Evaluating Temporal Misalignment}\label{sec:time-eval}
Prior work has shown that a dataset's collection time can affect the downstream model's abilities \citep{lazaridou2021mind,agarwal2022temporal}.
\citet{luu2021time} release several datasets in which increasing temporal distance between \textit{finetuning} and evaluation time decreases test performance.
We choose 5 of these datasets from varying domains to evaluate whether a similar phenomenon exists between \textit{pretraining} and evaluation time: PubCLS, NewSum, PoliAffs, TwiERC, and AIC.

\vspace{-3mm}
\paragraph{Evaluating Toxic Generation}
Generateing profane, sexually explicit, insulting, or obscene text or text that attacks identity groups or targets protected human attributes limits the applications LMs may be used for  \citep{gehman2020realtoxicityprompts}.
% LMs cannot be used for many important applications if they generate profane, sexually explicit, insulting, or obscene text or text that attacks identity groups or targets protected human attributes \citep{gehman2020realtoxicityprompts}.
We evaluate this behavior with language model prompts designed to elicit biased or toxic outputs related to gender, race, and religion \citep{chowdhery2022palm}, and then measuring the fraction of generated continuations which are assigned a high toxicity score by the Perspective API (see Appendix \ref{app:toxicity-details} for details).
We also use the RealToxicityPrompts dataset \citep{gehman2020realtoxicityprompts}, which consists of text excerpts from the OpenWebText dataset \citep{Gokaslan2019OpenWeb} that were labeled as toxic by the Perspective API.

\vspace{-3mm}
\paragraph{Evaluating Toxicity Identification}
\label{sec:eval-tox-identification}
While some applications require LMs not to generate toxic text, in other applications it is important for LMs to \textit{recognize} such language.
\emph{Toxicity Identification} has become particularly critical as a step in content moderation for major communication platforms \citep{nyt2020moderation, newamerica2019moderation}.
Definitions vary by setting, targeting hate speech, stereotypes, social bias, or some definition of toxicity.
We evaluate this ability with a variety of toxicity interpretations, using train and test sets from Social Bias Frames (SBF, \citealp{sap-etal-2020-social}), DynaHate (DH, \citealp{vidgen-etal-2021-learning}), and Toxigen \citep{hartvigsen2022toxigen}.\footnote{We use the offensiveness detection task from Social Bias Frames. DynaHate releases 4 rounds of adversarial datasets, for which we use the test sets for Round 3 (R3) and Round 4 (R4).}

\vspace{-3mm}
\subsection{Models}

For all our experiments, we use two sizes of decoder-only, Transformer-based language models, trained in the T5X codebase \citep{roberts2022t5x}.
Our main experiments use \bigLM, a 1.5B parameter decoder-only model similar to the t5.1.1-XL architecture configuration  trained with an autoregressive next-token-prediction objective.
For experiments that measure scaling effects, we use \smalLM, a 20M parameter decoder-only model similar to the t5.1.1-small configuration.
These configurations are popular, show  decent performance \citep{wang2022language} and can generate text without additional finetuning.
Additional details on pretraining and finetuning are available in \cref{sec:app-exp-details}

\vspace{-3mm}
\section{Impact of Data Curation on Data Characteristics}
\label{sec:data-analysis}

% https://www.latex4technics.com/?note=MOB
\begin{tcolorbox}[width=\textwidth,title={Section Findings}] % ,colbacktitle=yellow,coltitle=blue     ,colback={green}
\vspace{-2mm}
% \begin{itemize}[noitemsep, topsep=0pt]
\begin{itemize}[itemsep=0pt, wide=3pt]
    \item The Pile's documents are on average longer, more readable and higher quality than documents in C4 but contain more personally identifiable information (PII).
    \item Books is an outlier domain, having the longest, most readable, most toxic, and most PII-filled documents, 
    % while also being high quality.
    while also containing high-quality text.
    \item High toxicity and low quality documents have similarly high PII amounts but otherwise
    % opposite mean characteristics in
    have very different average
    length and quality and toxicity levels.
    \item More recent web-scraped text is 
    % Over time, C4 text becomes 
    more diverse and less toxic but also lower quality.
\end{itemize}
\end{tcolorbox}

Before evaluating the effect of data ablations on models, we present observational statistics on the pretraining datasets themselves.
This analysis reveals how the Pile's domains compare to C4 and to one another, and how curation or filtering choices impact features of the data, sometimes inadvertently.
We find that there are substantial interactions between curation choices.

We calculate a range of features for each document, including toxicity and quality metrics; categories of personally identifiable information (PII); and text statistics such as average word length, readability, type-token ratio, and sentiment. For more details and analysis on these features see \cref{sec:app-data-analysis}.

\begin{figure}[ht]
    \centering
        \subfigure[Domains]{
        \includegraphics[height=.65\linewidth]{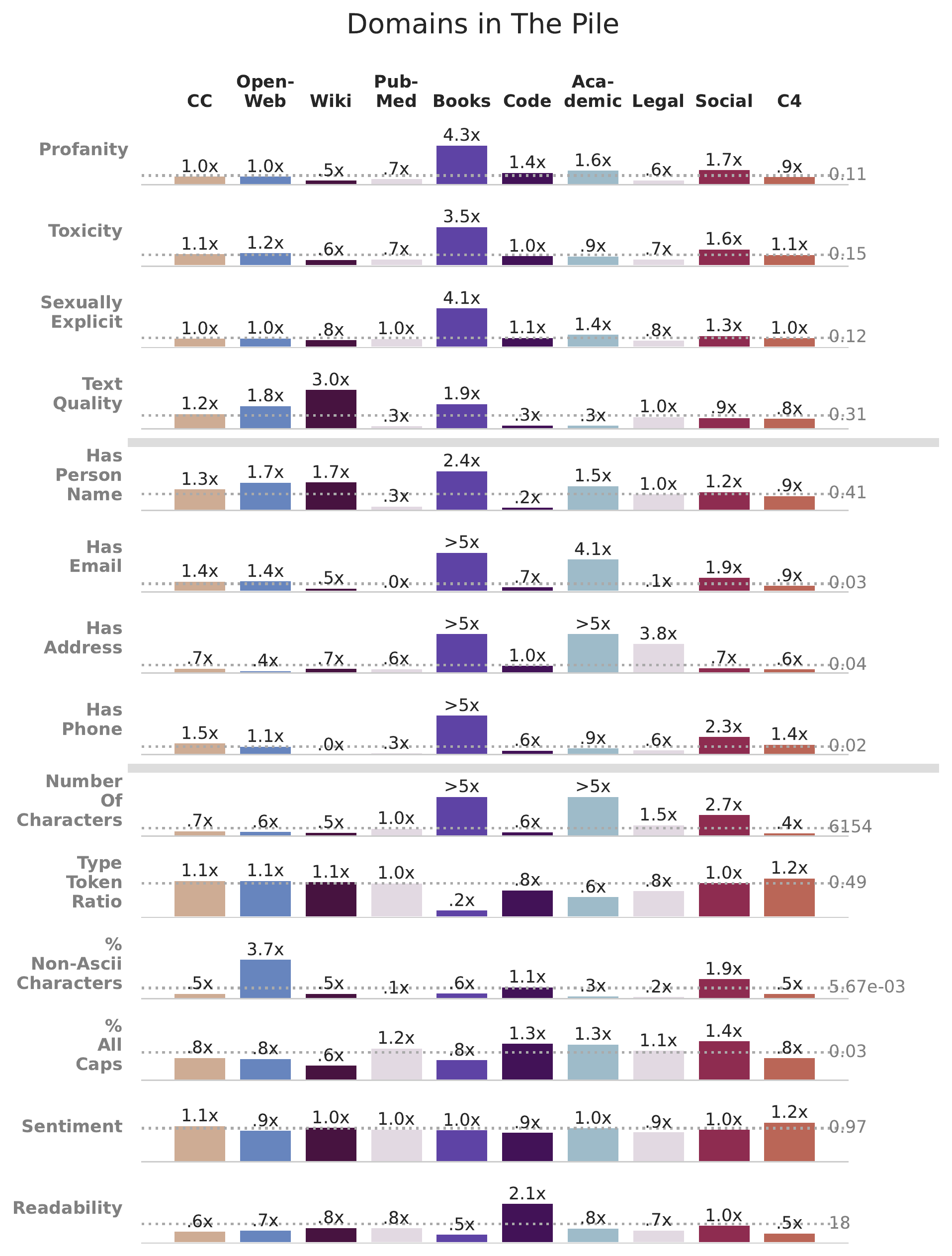}
    }
    \subfigure[Toxicity/quality in C4]{
        \includegraphics[height=.65\linewidth]{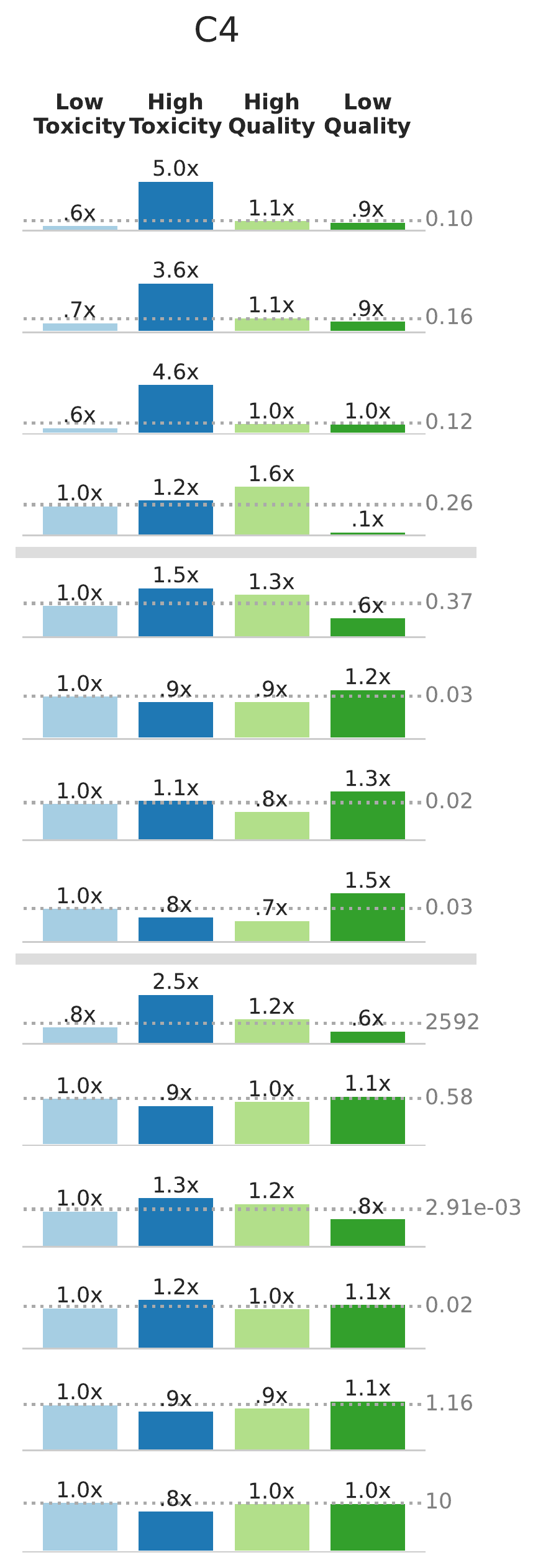}
    }
    \subfigure[Toxicity/quality in the Pile]{
        \includegraphics[height=.65\linewidth]{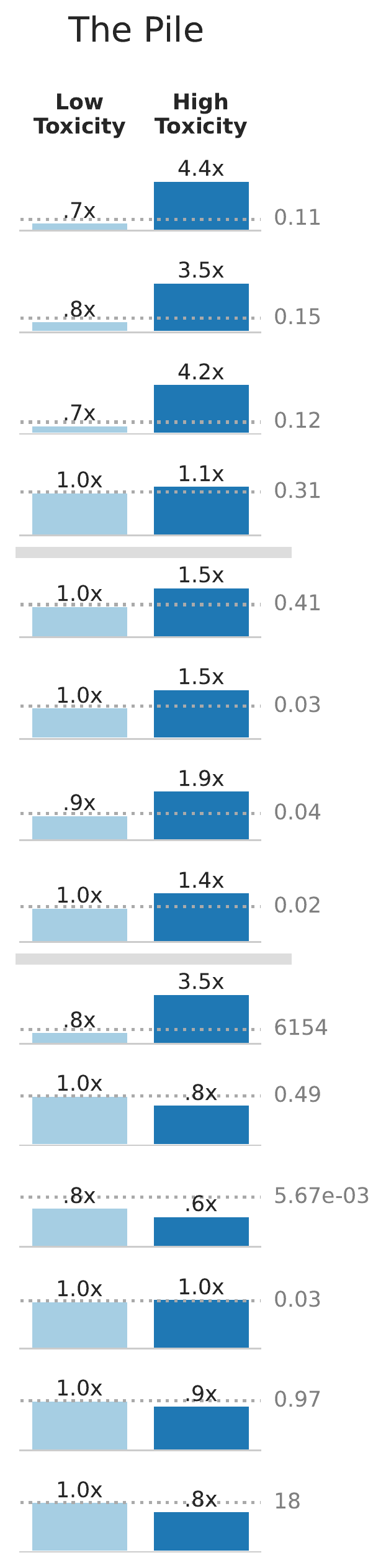}
    }
    \caption{
     \small \textbf{Feature differences across slices of the pretraining datasets.} 
     Bars show the ratio between the mean feature value for the slice and the mean value for the dataset (the Pile or C4), which is indicated by a horizontal gray line.
     For example, Wiki text has half the \textit{profanity} and three times the \textit{quality} values as the average for the Pile.
     %Bar height indicates the mean feature value, annotated with the ratio beween the dataset and the baseline, which is set to the Pile or C4.
     %The gray dashed line and gray number show the baseline's actual value.
    }
    \vspace{-3mm}
    \label{fig:composition-analysis}
\end{figure}

\vspace{-3mm}
\paragraph{C4 vs the Pile}
\cref{fig:c4-pile-composition} shows the differences between the two source datasets. 
Documents in the Pile are on average longer (2.4x), have more non-ASCII characters (1.9x) indicating greater linguistic range, and are also measured as higher quality (1.2x) and more readable (1.8x).
Pile documents also contain more PII, in particular personal names, addresses, and emails.

\vspace{-3mm}
\paragraph{Toxicity and Quality}
While it is reasonable to assume that high toxicity should correlate with low quality, 
\cref{fig:composition-analysis} shows that the relationship is more complicated: 
in fact, toxicity and quality are not well-aligned with one another.
High toxicity documents have higher text quality than low toxicity documents.
There is also little discernible difference in feature measurements for profanity, toxicity, and sexually explicit content between content classified as low vs. high quality.

\vspace{-3mm}
\paragraph{Domains}
Looking at characteristics of the Pile by domain in \cref{fig:composition-analysis} suggests an explanation.
The Books subset stands out as having substantially more profane, toxic, and sexual content, but also greater predicted quality.
While we might expect books to be high quality, 
% While books represent quality 
in the sense that they typically contain meaningful, well-edited sentences, they also contain strong language and erotic subjects.
This may also explain why documents classified as high toxicity in both C4 and the Pile are much longer (2.5x and 3.5x respectively), more profane (5x and 4.4x), sexually explicit (4.6x and 4.2x), and toxic (3.6x and 3.5x).
However, Pile documents with high toxicity are 1.4-1.9 times more likely to have PII of various kinds, while in C4 this is not true.
Documents classified as high quality in C4 were longer (1.3x and 1.2x), and had more names (1.6x and 1.8x), but fewer emails, addresses, and phone numbers.

Among the domains we studied, OpenWeb provides the most lexical and linguistic diversity, with the highest non-ASCII characters and type-token ratio.
Wikipedia presents the highest quality text, before Books and OpenWeb.
Technical domains such as PubMed, Code, and Academic score low on predicted quality, indicating that overly-specific positively-defined filters on web documents may remove substantial amounts of potentially useful specialized text.

\vspace{-3mm}
\paragraph{Time} Comparing across different collection times of C4 (in \cref{fig:c4-pile-composition}), we see a couple of steady trends.
The percentage of non-ASCII characters increased steadily in more recent years while the measured text quality declines.
This growth may be due to increasing non-English content, but could also correspond to rising use of emojis and non-ASCII punctuation.
Toxicity scores also decrease slightly in later years, while sentiment increases. 

% \vspace{-3mm}
% \section{Impact of Data Curation on Pretrained Models}

\vspace{-3mm}
\section{Impact of Dataset Age on Pretrained Models}
\label{sec:time}

% https://www.latex4technics.com/?note=MOB
\begin{tcolorbox}[width=\textwidth,title={Section Findings}] % ,colbacktitle=yellow,coltitle=blue     ,colback={green}
\vspace{-2mm}
% \begin{itemize}[noitemsep, topsep=0pt]
\begin{itemize}[itemsep=0pt, wide=3pt]
    \item Both models and evaluation datasets become stale.
    \item Temporal misalignment between pretraining and evaluation data is not overcome by finetuning.%Differences in pretraining and evaluation year degrade performance and are not overcome by finetuning.
    \item
    % This phenomenon complicates NLP experiments that compare old and new models on a given evaluation dataset.
    Temporal misalignment complicates evaluation of models trained at different times, as older evaluation datasets may become stale and newer evaluation datasets may under-estimate performance of older models. 
    % \item This phenomenon may complicate NLP experiments, %e.g. when comparing old and new models on new data, 
    % as older evaluation datasets may become stale and newer evaluation datasets may under-estimate performance of older models. 
    % we may over-estimate the difficulty of newer evaluation sets benchmarked on older models.
    \item The effects of pretraining misalignment are stronger for larger models than smaller models. %XL models than Small models.
\end{itemize}
\end{tcolorbox}

While models are frequently and cheaply updated with new finetuning data, the expense of pretraining means the NLP community has relied on relatively few static pretrained models that are rarely updated or exchanged.
BERT, RoBERTa, GPT-2, and T5 variants, all pretrained prior to 2020, constitute the majority (estimated at \textasciitilde58\% as of April 16, 2023) of all models downloaded on HuggingFace.
Prior work demonstrates that language use changes over time \citep{altmann2009beyond,labov2011principles} and that \emph{temporal misalignment} between finetuning and evaluation datasets correlates with degraded performance, visible across settings and domains \citep{luu2021time,lazaridou2021mind,agarwal2022temporal,jang2022temporalwiki}.
In contrast, we examine the effect of temporal misalignment between \emph{pretraining data} and evaluation.
In evaluating the impact of pretraining time across data domains, we can quantify the impact this design choice has on NLP broadly.

\begin{figure}[ht]
    \centering
    \includegraphics[width=0.99\linewidth]{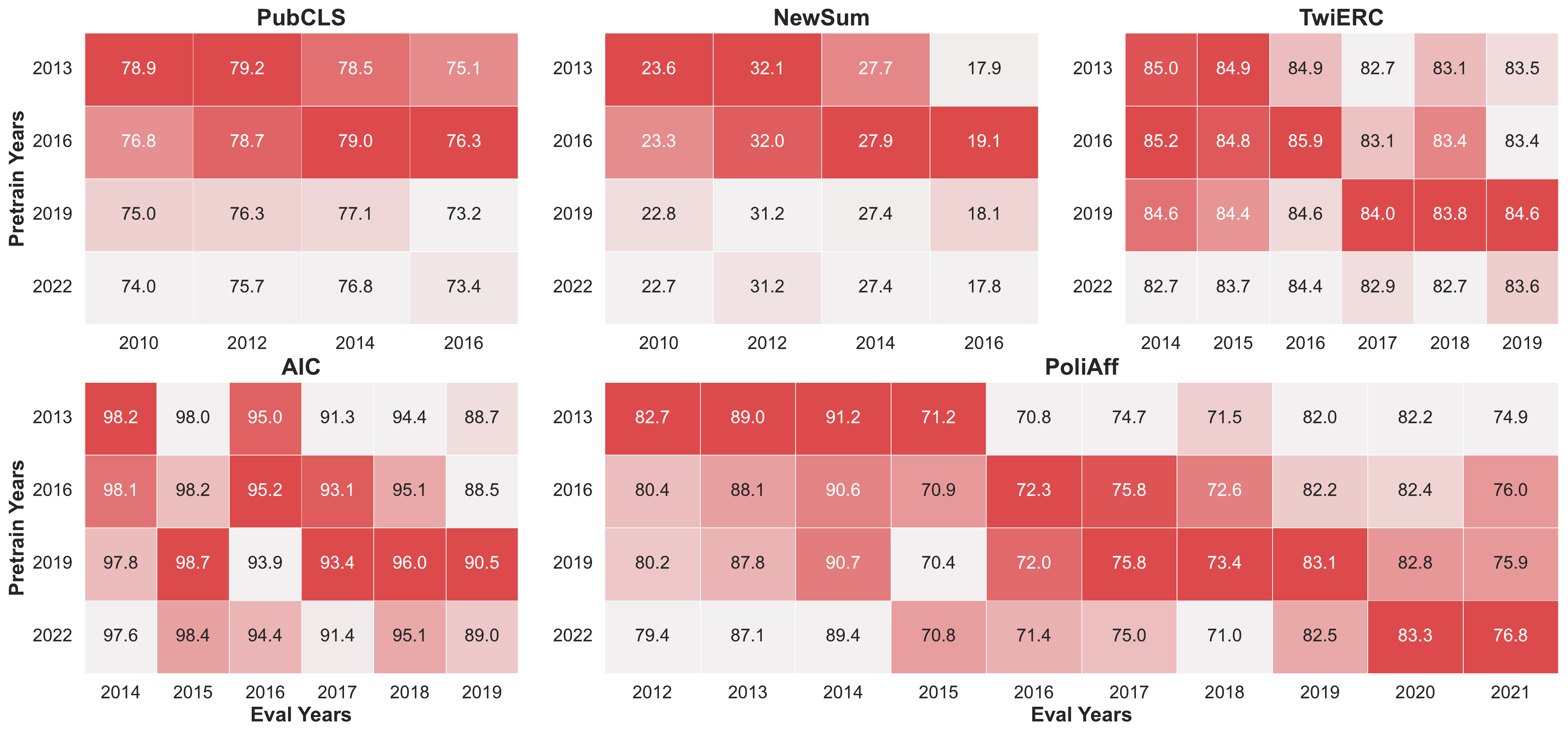}
    \caption{
    \small
    \textbf{Temporal Misalignment between Pretraining and Evaluation causes performance degradation}. 
    Four \bigLM's, each pretained on a different C4 time split, are evaluated on each time split across five datasets.
    Heatmap colors are normalized by column, following \citet{luu2021time} to show the best pretraining year for each evaluation year.
    }
    \vspace{-3mm}
    \label{fig:time-heatmap-xl-pt}
\end{figure}

We pretrain four autoregressive language models on versions of C4: 2013, 2016, 2019, and 2022.
For each version we begin with Common Crawl data and remove all data that was scraped after the cutoff year.
%Each version includes all Common Crawl data available from and before their respective scrape time (but not after).
Following \citet{luu2021time}, we measure the effect of temporal misalignment by using evaluation tasks (from News, Twitter, and Science domains) that have training and test sets split by year.
After pretraining, we finetune each model on each dataset's training-year split separately, then evaluate on every test-year split.
Full details and results are in \cref{app:time-eval-details} and \cref{app:time-results}, respectively.

First, we replicate the performance degradation observed by \citet{luu2021time} due to finetuning and evaluation misalignment on the five tasks in \cref{fig:time-heatmap-xl-ft}.
Next, we estimate the effects of temporal misalignment between \emph{pretraining} and evaluation (\cref{fig:time-heatmap-xl-pt}). 
Since all models were finetuned on the training sets of the evaluation tasks, we show that temporal misalignment during pretraining persists even with temporally-relevant finetuning data.

\begin{figure}[ht]
    \centering
    \includegraphics[width=0.6\linewidth]{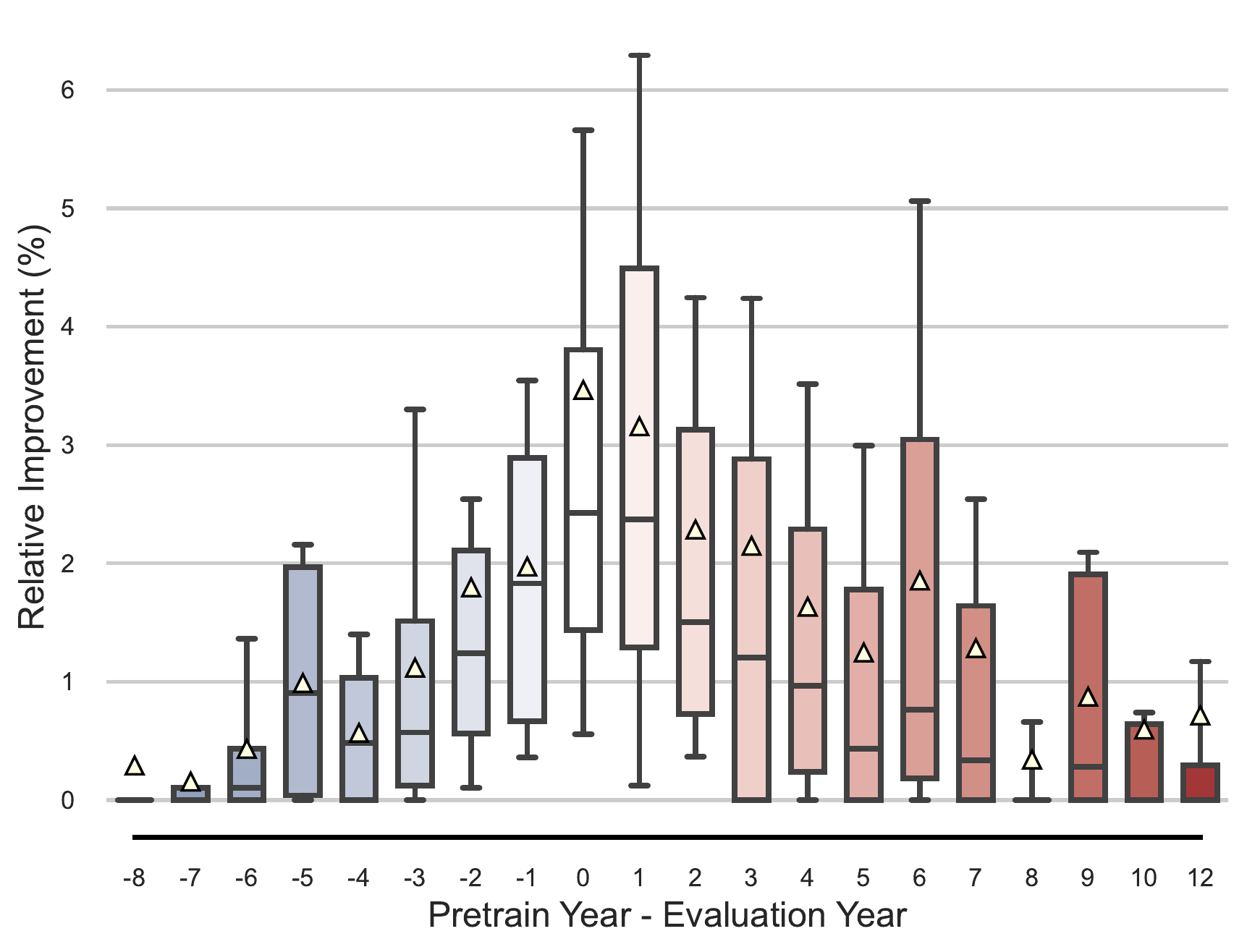}
    \caption{
    \small
    \textbf{The mean relative performance over 5 datasets (y-axis) increases as temporal misalignment (x-axis) approaches zero.}
    The boxplot indicates the median (solid line), mean (triangles), quartile range (boxes), and rest of the distribution (whiskers).
    Note that each dataset has different evaluation year ranges.
    }
    \vspace{-3mm}
    \label{fig:temporal-degradation-xl}
\end{figure}

\vspace{-3mm}
\paragraph{Performance degradation strongly correlates with pretraining misalignment and its effects are non-trivial.}
\citet{luu2021time} formalize a definition for Temporal Degradation (TD), which measures the performance change observed from one year difference between the finetuning and evaluation years.
We generalize TD to also measure the effect of one year difference between pretraining time and evaluation time, as described in \cref{app:temp-deg-measure}.
Furthermore, we measure the Pearson correlation $r$ between the performance difference and the temporal difference to understand the strength of the correlation.
In \cref{tab:time-td} we find temporal degradation is highest for finetuning (2.8 on average), as expected, but also surprisingly high for one year of pretraining (0.4)---particularly for the News domain.
The average Pearson correlation of $0.61$ indicates a strong correlation between pretraining temporal misalignment and performance degradation.
All five tasks pass a one-sided Wald test with $p<0.05$, validating the slope is greater than zero.

\begingroup
\setlength{\tabcolsep}{4pt}
\begin{table*}[ht]
    \centering
    \small
    % \begin{tabular}{>{\raggedright}p{7cm} crll}
    \begin{tabular}{l | l | rr | rr || rr | rr}
    \toprule
    & &  \multicolumn{4}{c||}{\textsc{Finetuning}}   & \multicolumn{4}{c}{\textsc{Pretraining}} \\
    & & \multicolumn{2}{c|}{\textsc{ \smalLM{} }} & \multicolumn{2}{c||}{\textsc{\bigLM{}}} & \multicolumn{2}{c|}{\textsc{\smalLM{}}} & \multicolumn{2}{c}{\textsc{\bigLM{}}} \\
    \textsc{Domain} & \textsc{Task} & \textsc{TD} & \emph{r} & \textsc{TD} & \emph{r} & \textsc{TD} & \emph{r} & \textsc{TD} & \emph{r} \\
    \midrule
    \multirow{2}{*}{\textsc{News}} & \textsc{PubCLS} & 5.82 & 0.84 & 5.63 & 0.80 & 0.02 & 0.01\textsuperscript{\textdagger} & 0.59 & 0.67 \\
    & \textsc{NewSum} & 0.80 & 0.82 & 2.91 & 0.92 & -0.31 & -0.29 & 0.73 & 0.45 \\
    % \midrule
    \multirow{2}{*}{\textsc{Twitter}} & \textsc{PoliAff} & 3.74 & 0.84 & 4.93 & 0.89 & 0.50 & 0.21 & 0.28 & 0.56 \\
    & \textsc{TwiERC} & 0.49 & 0.73 & 0.53 & 0.82 & 0.05 & 0.27 & 0.23 & 0.72 \\
    % \midrule
    \textsc{Science} & \textsc{AIC} & 0.94 & 0.83 & 0.24 & 0.36 & 0.11 & 0.18\textsuperscript{\textdagger} & 0.23 & 0.66 \\
    \bottomrule
    & \textsc{Mean} & 2.36 & 0.81 & 2.84 & 0.76 & 0.08 & 0.07 & 0.41 & 0.61 \\
    \bottomrule
    \end{tabular}
    \caption{
    Temporal Degradation (TD) measures the expected performance degradation from one year of temporal misalignment.
    We report TD first between finetuning and evaluation, then pretraining and evaluation, for \bigLM{} and \smalLM{}, across five tasks.
    Pearson correlation $r$ indicates the correlation strength between performance and temporal change.
    \textbf{Temporal Degradation due to pretraining is significant and persistent across domains.}
    All correlations are significant at $p<0.05$ unless marked with \textsuperscript{\textdagger}.}
    \label{tab:time-td}
\end{table*}
\endgroup

\vspace{-3mm}
\paragraph{Pretraining misalignment is not overcome by significant finetuning.}
The temporal degradation due to pretraining suggests models pretrained on data from the same time frame as target evaluations will have advantages over models trained on much older or newer data.
Notably, this effect is observed for models which are finetuned on the full temporally-relevant training sets.
This suggests that even substantial finetuning cannot overcome pretraining data that is temporally misaligned.

\vspace{-3mm}
\paragraph{Pretraining misalignment effects are asymmetric and have implications for NLP evaluations.}
We observe performance degradation regardless of whether the pretraining data was collected before or after the evaluation data.
While we would not expect a 2019 checkpoint to perform well on questions about COVID, we also find that 2022 checkpoints perform less well on Obama-era evaluations than earlier models.
In particular, \cref{fig:temporal-degradation-xl} shows performance degradation is asymmetric: it is steeper when the evaluation year is after the pretraining year (blue bars) as opposed to the reverse (red bars).
This finding suggests that both models and evaluations become stale: older models perform less well than newer models on new evaluations and newer models will perform less well on older evaluations.
This phenomenon may have subtle implications for NLP experiments comparing models pretrained at different times.
For instance, newer evaluation sets may appear much more difficult than old evaluation sets when applied to established, but less fresh, models.
Similarly, older evaluations may underestimate the capabilities of newer models.

\vspace{-3mm}
\paragraph{Temporal Degradation is greater for larger models}
We find more temporal degradation for \bigLM{} (1.5B parameters) than for \smalLM{} (20M parameters). 
As shown in \cref{tab:time-td}, we do not find the same temporal degradation effects of pretraining were significant for \smalLM{} models.
This suggests that larger models may have a greater sensitivity to temporal information than smaller models, which may not have the capacity to take advantage of subtle temporal features at all.
Full results for \smalLM{} experiments are provided in \cref{app:time-results}.

\vspace{-3mm}
\section{Impact of Quality \& Toxicity Filters on Pretrained Models}
\label{sec:quality-toxicity}

% https://www.latex4technics.com/?note=MOB
\begin{tcolorbox}[width=\textwidth,title={Section Findings}] % ,colbacktitle=yellow,coltitle=blue     ,colback={green}
\vspace{-2mm}
% \begin{itemize}[noitemsep, topsep=0pt]
\begin{itemize}[itemsep=0pt, wide=3pt]
    \item Quality and toxicity filters have very different effects.
    \item Quality filters improve performance significantly, despite removing training data.
    \item Quality filtering effects are not easily predicted by dataset characteristics. Future filters should weigh more than one dimension of quality.
    \item Toxicity filtering trades off generalization and toxicity identification ability for reduced risk of toxic generation.
    % ---one size does not fit all.
    \item When optimizing for toxicity identification tasks, practitioners should use an inverse toxicity filter.
\end{itemize}
\end{tcolorbox}

Most modern large language models use some form of quality and/or toxicity filtering for their pretraining datasets (\cref{tab:model_data_survey}).
To curb toxicity, T5 uses $n$-gram filters, Gopher and Chinchilla use SafeSearch filters, and LaMDA uses ``safety discriminators''.
Quality heuristics are universally applied for web-scraped data, with newer models like LLaMA, the GPT-series and the PaLM-series all relying on quality classifiers.
% Despite their importance, to our knowledge no prior work has published empirical comparisons of these filter types.
To compare and quantify the effects of these two filter types, we implement quality and toxicity filters at various thresholds, as described in \cref{sec:data-filters}, to vary the quantity of toxic and low-quality text present when pretraining models on the Pile and C4.

\begin{figure}[t]
    \centering
    
    \subfigure{
        \includegraphics[width=.45\textwidth]{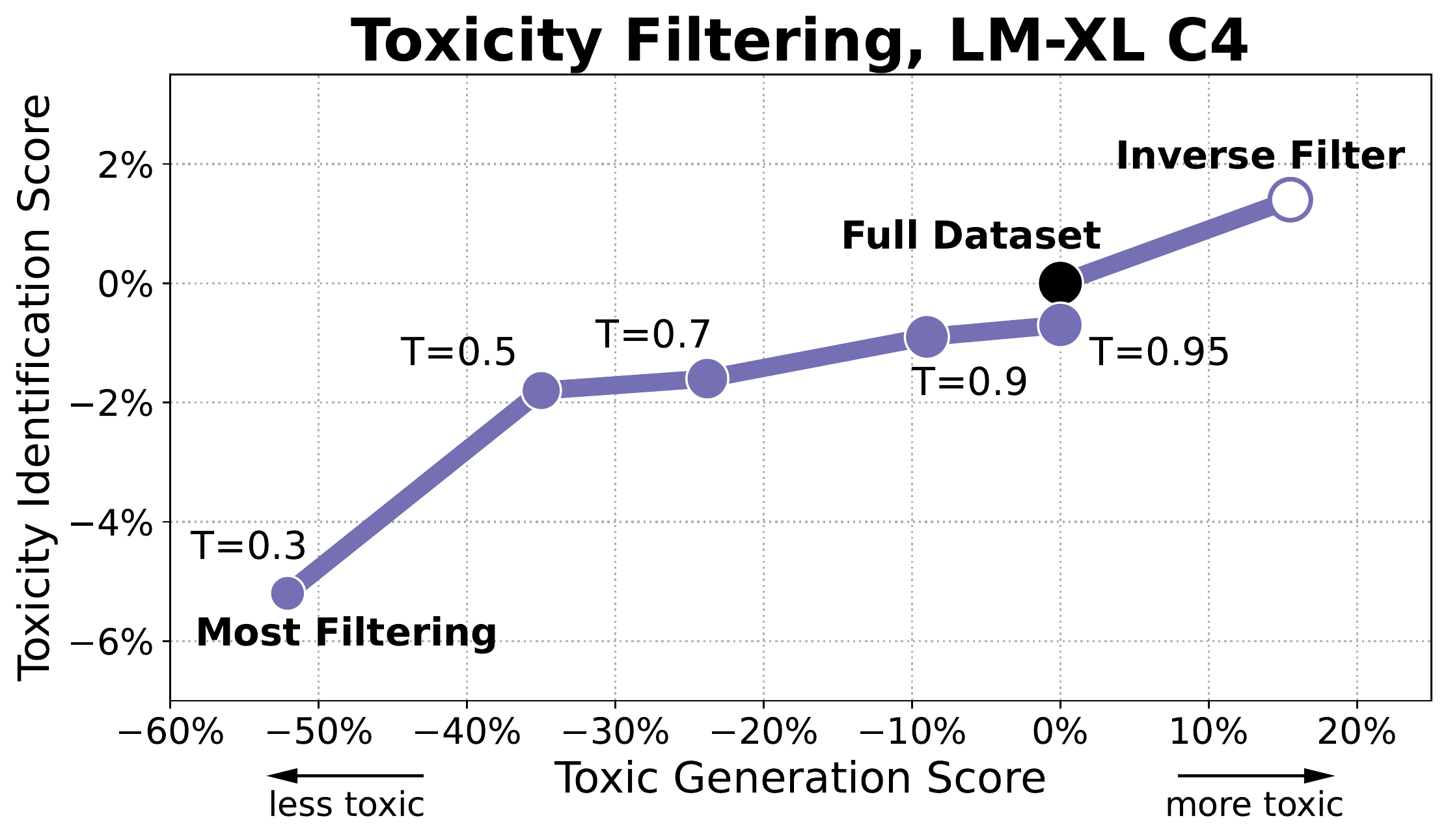}
    }
    \subfigure{
        \includegraphics[width=.45\textwidth]{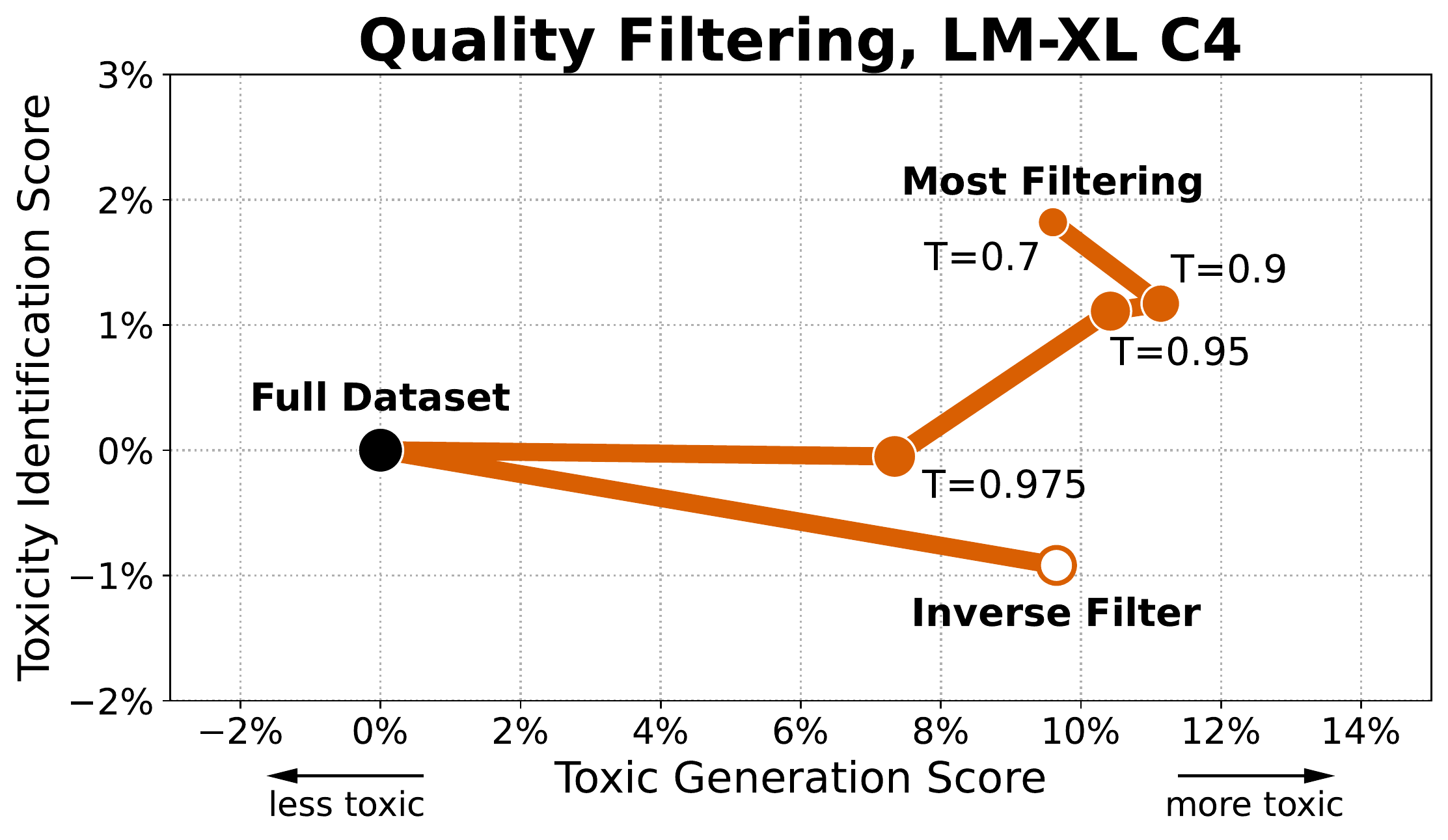}
    }
    \caption{
    \textbf{Toxicity filtering the pretraining dataset decreases the ability of \bigLM to identify toxicity and to generate toxic text. Quality filtering surprisingly increases both abilities.} Documents with scores below a given threshold were filtered out.
    }
    \label{fig:tox_qual_filter_c4}
    \vspace{-3mm}
\end{figure}

\begin{figure}[t]
    \centering
    \includegraphics[width=.98\textwidth]{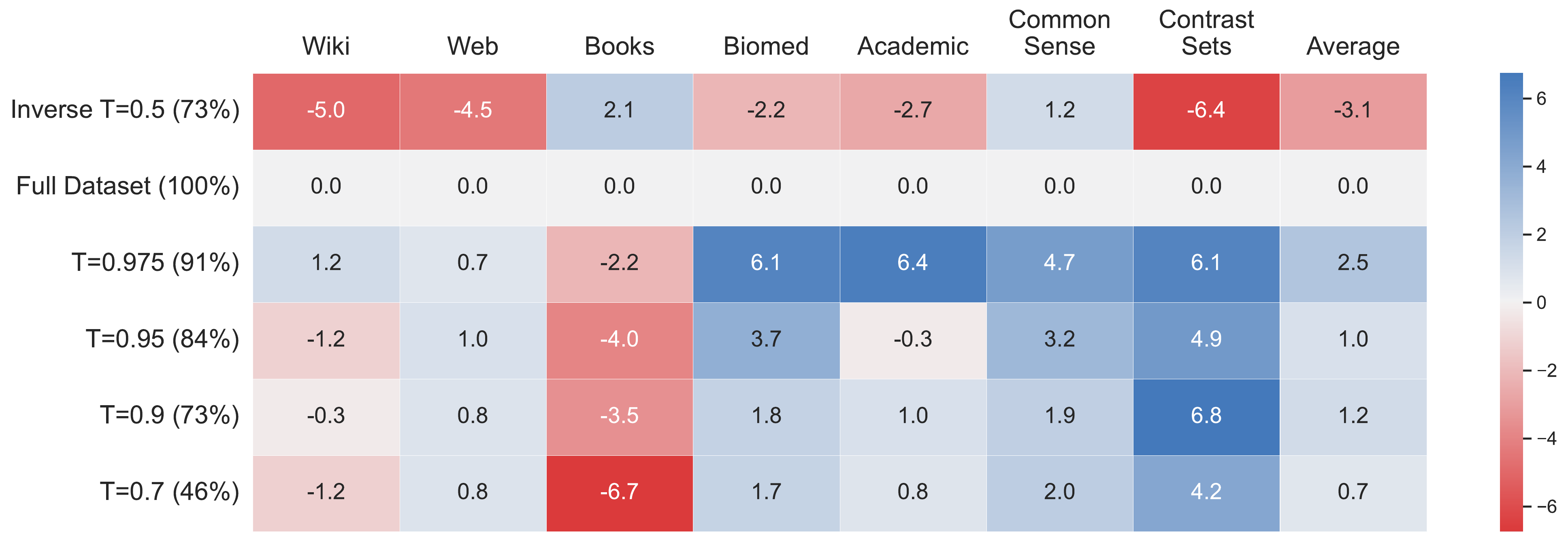}
    \caption{
    \textbf{Quality filtering C4 increases \bigLM's downstream performance on all QA task domains, except for Books.} 
    The quality filter threshold is on the x-axis, with percentage of training data remaining in parenthesis. 
    Each column represents a set of QA evaluations from a domain.
    The `Full Dataset' is unfiltered, and the `Inverse' filter removes the highest quality data instead.
    }
    \label{fig:qual_filter_on_domains}
\end{figure}

\begin{figure}[t]
    \centering
    \includegraphics[width=.98\textwidth]{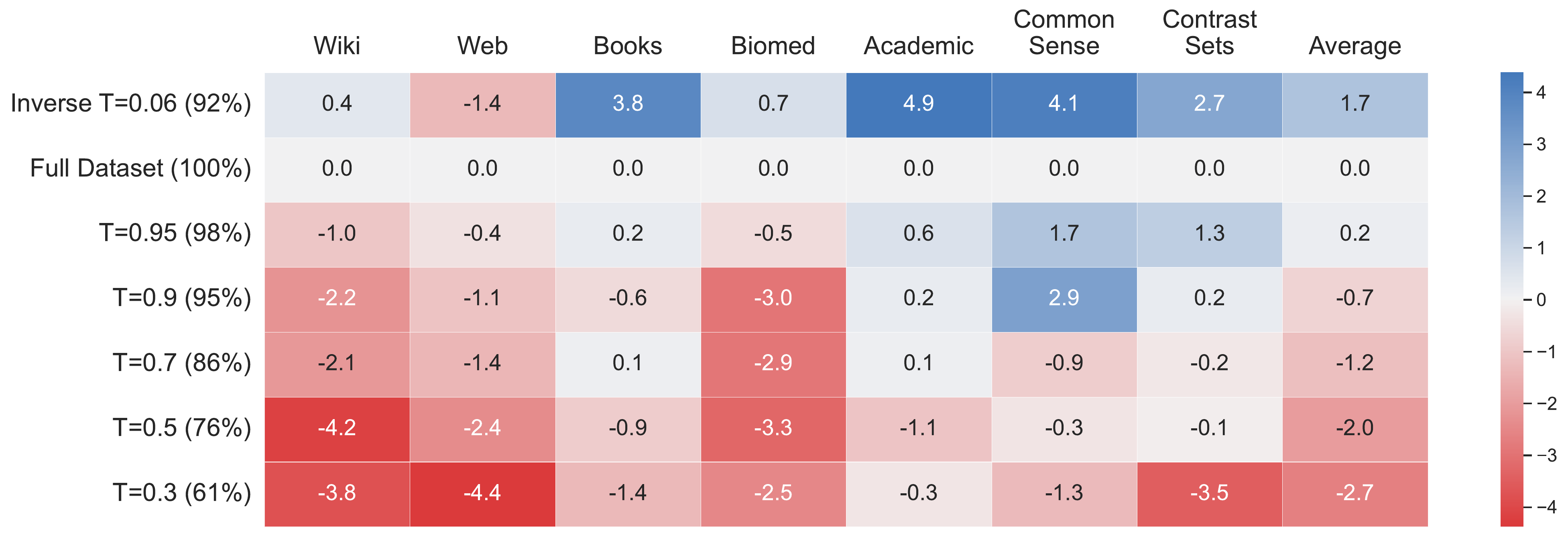}
    \caption{
    \textbf{Toxicity filtering C4 reduces \bigLM's downstream performance on most QA task domains.}
    The toxicity filter threshold is on the x-axis, with percentage of training data remaining in parentheses. 
    Each column represents a set of QA evaluations from a domain.
    The `Full Dataset' is unfiltered, and the `Inverse' filter removes the lowest toxicity data instead.
    }
    \label{fig:tox_filter_on_domains}
\end{figure}

\vspace{-3mm}
\paragraph{Quality filters significantly improve performance across nearly all tasks, despite reducing training data quantity and variety.}
We see the quality filters improve nearly all downstream tasks: toxicity identification by 2\% (\cref{fig:tox_qual_filter_c4}, right) and most QA task categories by 1-6\% (\cref{fig:qual_filter_on_domains}).
Of most interest, these improvements are realized despite removing 10\%+ of the training data, even though we find that removing data usually leads to a decrease in performance (\cref{sec:domain-comp}).
While the average performance peaks at $T=0.975$ for the QA tasks, greater quality filtering still outperforms the unfiltered baseline on average.
For the toxicity identification experiments, the performance is still improving after $T=0.7$, where 55\% of the dataset has been filtered out.

\vspace{-3mm}
\paragraph{Dataset quality characteristics are not strongly indicative of filtering effects.}
In \cref{sec:data-analysis}, Books, Wikipedia, and Web data are classified as highest quality.
\cref{fig:qual_filter_on_domains} shows that despite this, quality filtering provides the least benefit to QA tasks in these categories, even hurting the performance for Books.
On the other end, academic and biomedical data are ranked among the lowest quality, but their QA tasks benefit the most from quality filtering.

\vspace{-3mm}
\paragraph{Optimizing on one measure of quality is not sufficient to predict or improve performance across domains.}
Most interestingly, Wikipedia and Web QA tasks are among the most hurt by the inverse filter---suggesting these domains are not affected as much by the absence of the lowest quality data as the presence of the highest quality data.
Also unexpectedly, both the quality and inverse quality filters led to models with higher toxic generation tendencies (\cref{fig:tox_qual_filter_c4}, right)---the one dimensional measure of quality captured by the quality scores is not sufficient to explain this behaviour.
In other words, different segments of data along this classifier's quality spectrum can have strong but varied effects on different domains.
It suggests practitioners should move beyond one measurement of quality and consider multiple.

\vspace{-3mm}
\paragraph{One size does not fit all. Toxicity Filtering leads to a trade-off between toxic identification and toxic generation goals.}
Filtering using a toxicity classifier, we find a trade-off: models trained from heavily filtered pretraining datasets have the least toxic generation but also the worst toxicity identification (\cref{fig:tox_qual_filter_c4}, left).
Similarly, \cref{fig:tox_filter_on_domains} shows the performance of QA tasks unrelated to toxicity are hurt by toxicity filtering, though this may be due to the overall decrease in training data.
Ultimately, the intended behaviour of the model should inform the filtering strategy, rather than one size fits all.
Most interesting of all, the strongest performance on toxicity identification for every dataset comes from the inverse toxicity filter.
\textbf{Practitioners optimizing for performance on toxic domains should intentionally apply inverse filters}.
% User-facing text generation applications will call for lower toxic generation, but toxicity identification applications benefit instead from a large proportion of pretraining text with high toxicity.

\vspace{-3mm}
\section{Impact of Domain Composition on Pretrained Models}
\label{sec:domain-comp}

% https://www.latex4technics.com/?note=MOB
\begin{tcolorbox}[width=\textwidth,title={Section Findings}] % ,colbacktitle=yellow,coltitle=blue     ,colback={green}
\vspace{-2mm}
% \begin{itemize}[noitemsep, topsep=0pt]
\begin{itemize}[itemsep=0pt, wide=3pt]
    \item Inclusion of Common Crawl, OpenWeb and Books have the strongest positive effects on downstream performance.
    % , both on average and for targeted domains.
    Data source heterogeneity is more important than data quality or size.
    \item Targeted data %\emph{usually} 
    helps targeted evaluations, but not always as much as including heterogeneous web domains.  
    % \emily{maybe: This points to a need to better understand and document these large unstructured domains.}
    % \item The best performing models do not omit pretraining data sources.
    \item It is beneficial to include as many pretraining data sources as possible.
\end{itemize}
\end{tcolorbox}

As shown in \cref{tab:model_data_survey}, pretraining datasets seek to generalize to a wide array of downstream tasks by combining data from a diverse set of domains. 
How does the choice of pretraining source domains impact downstream performance?
We empirically answer this question by ablating pretraining sources from the Pile one-at-a-time and measuring the downstream performance change in 27 QA tasks from diverse domains.

\begin{figure}[ht]
    \centering
    \includegraphics[width=0.98\linewidth]{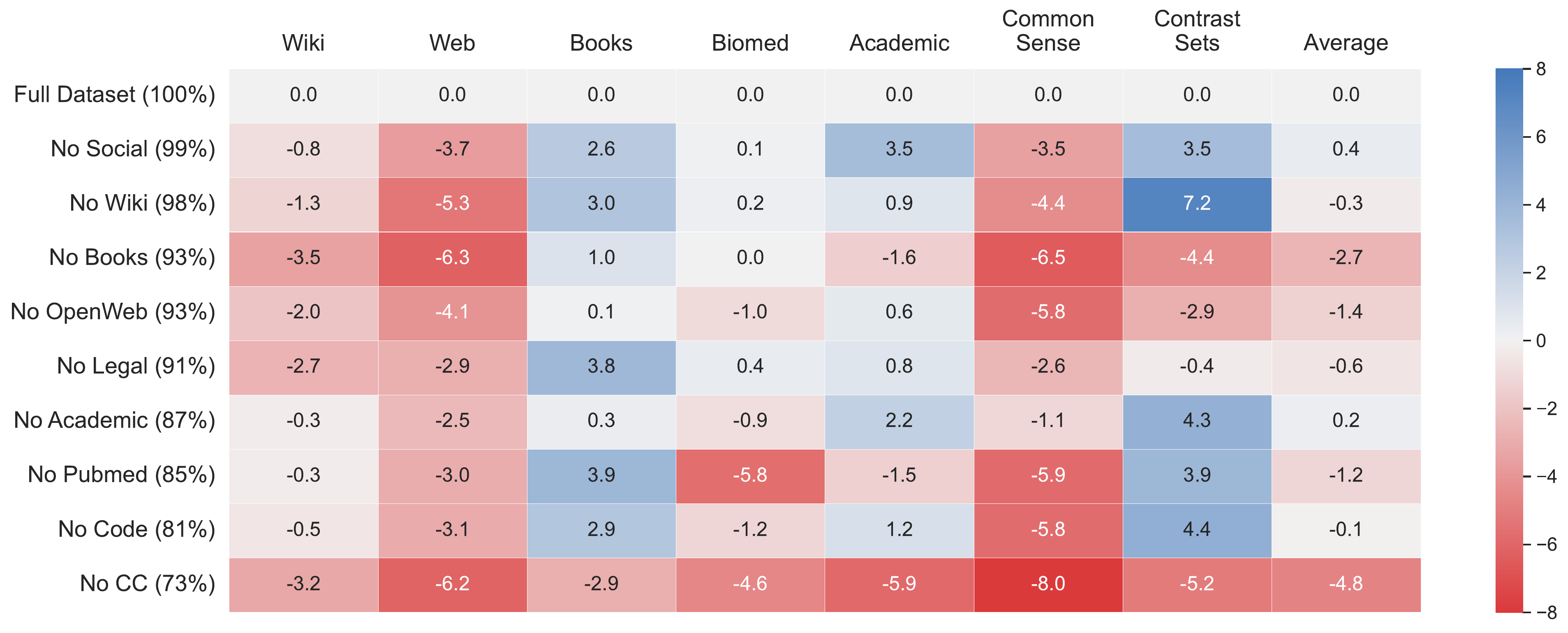}
    \caption{
    \small
    \textbf{QA tasks are affected by removing domains when pretraining \bigLM{}.} 
    Each row represents a model with one domain removed, the size of the remaining dataset is shown at the left in parentheses.
    Each column represents a set of QA evaluations from a domain.
    The \textsc{Full Dataset} model represents the unfiltered Pile \bigLM, and all scores are relative to this Base model. 
    %The percentage of remaining data after removing the slice is shown in the x-label parentheses.
    }
    \vspace{-3mm}
    \label{fig:domain-heatmap-xl}
\end{figure}

We first group the Pile data sources into nine domains representing conceptual sources that practitioners could choose to license or scrape more of: Common Crawl (CC), OpenWeb, Wikipedia, Books, PubMed, Academic, Code \& Math, Legal, and Social (see \cref{tab:pile-partitions}).
These are sorted in ascending order by size.
We choose to maintain the size disparities in these sources, simply because they reflect reality: curated Wikipedia content is innately finite, while web and books are much more abundant.
We then pretrain \bigLM{} with the full dataset minus each category, yielding nine models, then finetune each for QA using Natural Questions. 
Finally, we evaluate the model on 27 unique datasets from MRQA \citep{fisch2019mrqa} and UnifiedQA \citep{khashabi-etal-2020-unifiedqa} that have also been partitioned into domains. 
Full details are documented in \cref{app:domain-eval-details}.

\vspace{-3mm}
\paragraph{Common Crawl, OpenWeb, and Books have the strongest positive effects on downstream performance.}
\cref{fig:domain-heatmap-xl} shows that average downstream performance degrades the most when we remove web-based domains like CC, Books, and OpenWeb, corroborating recent findings by \citet{xie2023doremi}.
In particular, these sources improve performance on challenging Common Sense and Contrast Sets tasks.
While CC is the largest chunk of text in the Pile, Books and OpenWeb are smaller but provide the most heterogeneous and predicted-quality content (see \cref{sec:data-analysis}).
These results suggest that more data is not necessarily as important a factor as a combination of heterogeneity and quality.

\vspace{-3mm}
\paragraph{Domain heterogeneity is often more beneficial than targeted data, even for targeted evaluations.}
Ablating a pretraining domain has varying effects on downstream QA performance.
Predictably, performance degrades when we remove domains with close alignment between the pretraining and downstream data sources: removing PubMed hurts the BioMed QA evaluations, dropping Wikipedia hurts the Wikipedia benchmarks, and removing web content hurts web evaluations.
However, removing targeted domains does not necessarily have as significant an effect on related downstream domains as removing the large heterogeneous domains.
For instance, removing CC from the pretraining dataset reduces performance on downstream Academic QA tasks to a much greater extent than removing the Academic domain.
Our hypothesis is that CC, OpenWeb and Books contain extensive coverage of many topics, so removing the Academic-specific category of sources does not remove all relevant academic information.

\vspace{-3mm}
\paragraph{The best performing models use \emph{all} the pretraining data sources.}
Despite the importance of data heterogeneity, the best mean performance still comes from models that train on all, or nearly all, the data.
The exceptions are the removal of targeted source domains like the Pile's Code or Academic (advanced science and math journals) domains. These are both large but perhaps not well matched with the QA evaluation sets, which do not require coding skills or scientific rigour beyond that found on Wikipedia and from web-based sources.
This finding suggests that both the quantity and diversity of open source data remain a bottleneck for current pretraining methods. 

\vspace{-3mm}
\paragraph{Web and Books domains cause the biggest trade-off between toxic identification and generation.}
\label{sec:interaction-domain-toxqual}
We next consider whether reducing a model's pretraining exposure to toxic content affects either its propensity to generate toxic language or its ability to identify toxic language.
% In \cref{sec:domain-comp} we evaluated the effect of ablating pretraining domains on QA performance across domains.
%Here, we look at unintended effects of the inclusion of each domain on the ability of the downstream model to perform toxicity identification and generation.
\cref{tab:interaction-domain-tox-results} shows that the largest decreases in \textit{both} toxicity generation and identification were caused by removing CC (26.9\% of the data), OpenWeb (6.9\%), and Books (6.9\%).
This is consistent with the observation that Web and Books data had the highest concentration of text predicted to be toxic \cref{sec:data-analysis}.
% This is in accord with our finding in \cref{sec:interaction-toxqual-domain} that toxicity filtering reduces downstream QA performance in the Wiki and Web domains, suggesting that many documents in these domains might be classified as toxic.
These results suggest a trade-off: better performance on QA (\cref{sec:domain-comp}) and toxicity identification comes at the cost of more toxic generation.
% Unsurprisingly, removing Wikipedia (2.1\%) increased toxicity generation by 4.2\% above the non-filtered baseline model.
% Overall, none of these domain ablations led to the large drops in toxicity generation that we observed from the dedicated toxicity filters in \cref{sec:quality-toxicity}.

\begingroup
\setlength{\tabcolsep}{4pt}
\begin{table*}[t]
    \centering
    \small
    \caption{
    Effect of the Pile's domain composition on toxicity identification and generation.
    \textbf{Removing Books, CommonCrawl and OpenWeb lead to the greatest decrease in toxicity metrics. Removing Wikipedia had a strong increase in toxicity generation.}
    }
    % \begin{tabular}{>{\raggedright}p{7cm} crll}
    \begin{tabular}{l | r | cccc | r || ccc | r}
    \toprule
    \textsc{Filter} & \textsc{\% Data} & \multicolumn{5}{c||}{\textsc{Toxicity Identification ($\uparrow$)}} & \multicolumn{4}{c}{\textsc{Toxic Generation ($\downarrow$)}} \\
    & & SBF & Toxigen & DH R3 & DH R4 & Score & RTP-T & RTP-NT & RepBias & Score \\
    \midrule

    \textsc{Full Dataset} & 100.0 & 90.7 & 90.8 & 88.7 & 84.1 & 0.0 & 88.9 & 45.4 & 4.6$\pm$0.7 & 0.0 \\
    \textsc{No Social} & 98.8 & 90.9 & 91.0 & 87.8 & 84.9 & \cellcolor{forestgreen!1}+0.1 & 85.4 & 47.2 & 4.7$\pm$0.8 & \cellcolor{forestgreen!4}+0.4 \\
    \textsc{No Wiki} & 97.9 & 90.6 & 90.8 & 88.1 & 83.6 & \cellcolor{color3!4}-0.4 & 89.0 & 49.4 & 4.8$\pm$0.6 & \cellcolor{forestgreen!42}+4.2 \\
    \textsc{No Books} & 93.1 & 89.9 & 90.3 & 87.1 & 82.6 & \cellcolor{color3!13}-1.3 & 87.4 & 43.5 & 4.0$\pm$0.8 & \cellcolor{color3!62}-6.2 \\
    \textsc{No OpenWeb} & 93.1 & 89.9 & 90.3 & 86.4 & 82.5 & \cellcolor{color3!15}-1.5 & 88.0 & 42.1 & 4.3$\pm$0.6 & \cellcolor{color3!52}-5.2 \\
    \textsc{No Legal} & 91.0 & 90.9 & 90.8 & 88.1 & 83.0 & \cellcolor{color3!4}-0.4 & 88.2 & 46.1 & 4.7$\pm$0.8 & \cellcolor{forestgreen!8}+0.8 \\
    \textsc{No Academic} & 87.1 & 90.7 & 91.0 & 88.2 & 84.5 & \cellcolor{forestgreen!.5}+0.0 & 86.5 & 46.4 & 4.5$\pm$0.7 & \cellcolor{color3!12}-1.2 \\
    \textsc{No Pubmed} & 85.1 & 90.6 & 90.8 & 88.0 & 84.3 & \cellcolor{color3!2}-0.2 & 87.6 & 46.3 & 4.6$\pm$0.7 & \cellcolor{color3!2}-0.2 \\
    \textsc{No Code} & 80.9 & 91.0 & 91.2 & 88.5 & 84.5 & \cellcolor{forestgreen!2}+0.2 & 87.6 & 46.5 & 4.7$\pm$0.7 & \cellcolor{forestgreen!6}+0.6 \\
    \textsc{No CC} & 73.1 & 89.9 & 90.0 & 85.3 & 82.4 & \cellcolor{color3!19}-1.9 & 87.8 & 46.2 & 4.3$\pm$0.6 & \cellcolor{color3!21}-2.1 \\
    
    %  No Filter, Social, Wiki, Books, OpenWeb, Legal, Academic, Pubmed, Code, CC
    \bottomrule
    \end{tabular}
    \label{tab:interaction-domain-tox-results}
\end{table*}
\endgroup

\vspace{-3mm}
\section{Discussion}
\label{sec:discussion}

\vspace{-3mm}
\paragraph{Guided by intuition: undocumented \& unknown}
Pretraining dataset curation has been guided by intuitions: collections should be large, diverse, and high quality.
Decisions are often driven by the need for something ``good enough'' or by precedents that may themselves not have been thoroughly evaluated \citep{sambasivan2021everyone}.
Similarly, model developers occasionally neglect to share empirical insights, maintaining a knowledge gap, often referred to as ``documentation debt'' \citep{bandy2021addressing}.

Our results show that choices made in pretraining curation affect models in significant ways that cannot be easily erased by subsequent finetuning.
We urge both model producers and model users to think of dataset curation policies as a form of hyperparameter, much like learning rates or network dimensions.
Exhaustive search methods that work for single scalar values will not, however, scale to curation policies that affect terabytes of data.
While our results are necessary to establish that pretraining curation matters, they are not sufficient to answer all questions.
In this section we therefore make specific recommendations, but our primary result is that we need better tools for modeling the relationship between data and model capabilities.

\vspace{-3mm}
\paragraph{Age of the pretraining corpus.}
In an ideal world, models would be continuously re-trained on the most up-to-date data available.
However, given the expense of data collection and re-training, model creators must make a choice between efficiency and model staleness.
More subtly, we also find that using newer data can add a ``presentist'' bias when evaluating retrospective tasks.
% We investigate this tradeoff by recreating four versions of C4, each sources from a different scrape of the CommonCrawl, dating between 2013 and 2022.
The effect of staleness is not overcome even by plentiful finetuning data for the given task, and this effect is worse for the larger, more capable models.
%\kl{rephrase}
This result complements findings by \citet{schulman2023} that finetuning on newer data can aggravate hallucination for new data that is not well-grounded at pretraining time.
These tentative findings suggest the temporal properties of pretraining corpora are increasingly essential to consider for larger models, for more novel tasks (less finetuning data), and for instruction tuning models.
Current practice includes augmenting prompts with retrieved, recent data to help overcome stale pretraining data. While this can conceivably help mitigate staleness, retrieving relevant text is a challenge in its own right.

We recommend model creators report the temporal distribution of pretraining data, which is not currently standard practice \citep{hoffmann2022training, thoppilan2022lamda, anthropic2023claude, cohere2023}.
Users should be able to predict otherwise unforeseen performance degradations on much newer datasets, or be aware of the potential side effects of finetuning models on information not covered in pretraining.

\vspace{-3mm}
\paragraph{Data source composition.}
Decisions on the composition of a corpus intended for pretraining can have substantial impacts on downstream performance.
Of the two corpora we consider in this paper, C4 contains only one data source, a single scrape of the Common Crawl, while the Pile is a collection of 22 data sources.
It is more complex and costly to assemble a corpus which contains diverse sources, writing styles, and thematic areas.
Achieving this diversity might also leave models vulnerable to less careful curation or gaps in practitioner knowledge.

In our experiments, we ablate the Pile by systematically omitting each of its constituent datasets before pretraining, and then measuring the impact on standard benchmarks.
Our results suggest that practitioners should not omit any data sources if generalization to as many text-to-text tasks is the goal, and that future work should focus on collecting more diverse web and books content, which yield the largest benefits.
These findings are somewhat consistent with hypotheses that the volume of training data remains a limiting factor, especially given licensing constraints \citep{chinchilla2022implications}.

\vspace{-3mm}
\paragraph{Filtering for toxicity and quality.}
The Common Crawl contains an enormous amount of low quality (advertisements, repetitive, non-human-readable, etc.) and toxic text.
Many state-of-the-art language models filter out this text before training, either using bad words lists \citep{raffel2020exploring}, heuristics, or classifiers \citep{du_glam_2021,brown2020language,chowdhery2022palm}.
Deciding on how much and what kind of text to filter out requires non-trivial normative decisions, and all of these filtering approaches involve the model creator intentionally modifying the bias of their datasets and thus their models.

In our experiments, we expose an implicit trade-off between a model's generalization abilities and its tendency to generate toxic content.
This behavior is modulated by quality and toxicity filters.
In fact, over-sampling on \emph{more} toxic documents leads to the best performance on toxic identification.
This observation, coupled with evidence that recent work is using post-hoc methods to curb unwanted toxic generation (e.g. instruction tuning \citep{chung2022scaling} or steerable decoders \citep{dathathriplug, welbl2021challenges}), suggests practitioners should prioritize toxic identification rather than curbing toxic generation abilities during pretraining.
% Systems employing these additional techniques can prioritize toxic identification over toxic generation during pretraining---explaining the observed shift away from T5-style toxicity filters in proprietary models like PaLM, PaLM-2, and ChatGPT.

We find that our quality filter (the same used by PaLM, trained to keep content resembling Wikipedia and Books) significantly improves performance across domains, despite removing large portions of the training data.
Perplexingly, the Books domain is the one exception to the above observation, as its content ranks among the highest quality.
In general, observational quality characteristics of the data are not sufficient to predict which domains will benefit most from quality filtering.
Our analysis suggests that performance on a task/domain is not influenced \emph{only} by how much poor quality data (i.e. that which is unlike Wikipedia/Books) is removed, but also by other aspects of quality, such as how much of the highest or mid-quality data is represented along this specific measurement dimension.

\vspace{-3mm}
\section{Limitations}
\label{sec:limitations}

\vspace{-3mm}
\paragraph{Compute Expense \& Single Shot Experiments} To our knowledge, this is the largest publicly documented LM pretraining data ablation study, spanning 28 1.5B parameter models---training more models with different data variants from scratch than GLaM \citep{du_glam_2021}, miniBertas \citep{warstadt2020learning}, MultiBerts \citep{sellammultiberts}, and even Pythia \citep{biderman2023pythia}, which focuses on preserving data composition and order.
It is important to acknowledge each of these pretrainings, with their corresponding finetuning and evaluations is computationally and environmentally costly.
With this in mind, we made the careful decision on what experiments to pursue---narrowing our list to: age of the corpora, quality filters, toxicity filters, and the choice of source domains.
We carefully curated the choice of experiments in advance, without the luxury of multiple rounds of reflection and repetition, common in many NLP experimental settings.
As a result, we struck a balance as best we could between the computational costs, and reproducible validity.
We hope to justify the merits of our selection and also point out the surprises that motivate future work or a deeper look into the results.

\vspace{-3mm}
\paragraph{Blackbox APIs} An additional limitation is our use of Perspective's API  for evaluating the toxicity of generations.
While most of our toxicity filters and evaluations were in a compressed time period, \citet{pozzobon2023challenges} have since demonstrated the irreproducibility of black-box APIs, which may have shifting implementations over time.
We also believe that while this is the standard procedure for popular toxic generation benchmarks like RealToxicityPrompts, the reliance on APIs and narrow evaluation setting can have limited implications for toxic generation in real applications.
For the time being, these are the best proxies we have.

\vspace{-3mm}
\paragraph{English vs Multilingual Data}
Our analysis was limited to two English datasets. 
It’s important to note that training composition is an even more crucial question for multilingual and non-English models, where optimally balancing corpora from different languages and finding large-enough high-quality corpora can be very challenging \citep{chung2023unimax}.

\vspace{-3mm}
\paragraph{Relevance to Zero- \& Few-Shot Prompted Settings} Our experiments focus on finetuned settings rather than zero- or few-shot prompting.
This choice is motivated by finetuning being more applicable for 1.5B parameter models and also in many applied settings.
We cannot establish how well these findings translate to prompted settings (without finetuning), but suspect they are strong correlated.

\vspace{-3mm}
\section{Related Work}
\label{sec:rw}

\vspace{-3mm}
\paragraph{Pretraining Dataset Curation}
There have been dozens of general-purpose models trained for natural language understanding and generation tasks.
Early models in this space, such as ELMO \citep{peters-etal-2018-deep}, BERT \citep{devlin-etal-2019-bert}, and BERT's various descendants \citep{liu2019roberta,lan2020albert}, focused on strong finetuning performance for a variety of natural language inference tasks, as well as semantically meaningful language embeddings.
These systems were trained on semi-curated datasets such as Wikipedia, BookCorpus \citep{zhu2015aligning}, and news articles from the One Billion Word Benchmark \citep{chelba2013one}.
XLNet \citep{yang2019xlnet} broke away from this use of curated datasets to include documents from Common Crawl into their pretraining dataset.
T5 \citep{raffel2020exploring}, which introduced the C4 dataset, was one of the first pretrained language models to train exclusively on Common Crawl data.
Multilingual versions of T5 \citep{xue2021mt5} and BERT were trained on Common Crawl and Wikipedia, respectively.

GPT-2 was one of the first models intended primarily for generation \citep{radford2019language}.
Deeming Common Crawl too noisy to be practical for training generative models, they developed WebText, a dataset containing websites linked to from highly-ranked posts on Reddit.
Subsequent generative models proposed mixing large amounts of noisy Common Crawl data with smaller corpora perceived as high-quality.
The GPT-Neo model family \citep{black2022gpt} trained on the Pile, which augments the Common Crawl with ArXiV, Stack Exchange, legal documents, books, Github, and other more curated sourced \citep{gao2020pile}.
More recently, OPT \citep{zhang2022opt} trained on the Pile augmented with social media data \citep{baumgartner2020pushshift}, and LLaMA \citep{touvron2023llama} trained on C4 augmented with Github, Stack Exchange, books, and other sources. Pythia trained on the Pile, with and without duplication \citep{biderman2023pythia}.
Finally, the BLOOM model family \citep{scao2022bloom} trained on the ROOTS Corpus, which crowd-sourced a collection of ``identified'' datasets, coming from known, high-quality sources in a variety of languages.

All of the models mentioned so far are publicly available.
However, companies are increasingly training their best models on proprietary datasets, with only limited hints as to the data composition.
At Alphabet, models such as Gopher \citep{rae2021scaling}, GLaM \citep{du_glam_2021} , LaMDA \citep{thoppilan2022lamda}, and PaLM \citep{chowdhery2022palm} have been trained on mixtures of web text, books, news, code, Wikipedia, and dialog data.
At OpenAI, GPT-3 \citep{brown2020language} was trained on Common Crawl, WebText (GPT-2's training set), books, and Wikipedia.
Subsequent versions of their model have also included code.
Most of these models have acknowledged using various forms of filtering techniques to improve the quality of web-derived training data.
These include classifiers designed to exclude content which looks least like ``high-quality'' sources such as books or Wikipedia \citep{chowdhery2022palm,ouyang2022training}, using Google's SafeSearch for identifying toxic content \citep{rae2021scaling}, and various heuristics based on document length and the presence or absence of certain words or characters.

\vspace{-3mm}
\paragraph{Pretraining Dataset Analysis}
\citet{dodge2021documenting} find significant amounts of low-quality patent, military, and machine-generated text in C4, and a dearth of English text from American minority communities as well as from non-Western communities like India or Nigeria post-filtering, and so recommend against filtering.
In contrast, \citet{luccioni2021s} recommend more robust filtering practices to curb the significant presence of hate speech and sexually explicit content they find in C4 even after filtering.
Similarly, \citet{kreutzer2022quality} find that multilingual pretraining corpora are also dominated by low-quality text, particularly for lower resource languages.
\citet{lee2022deduplicating, kaddour2023minipile} show the benefits of deduplicating pretraining datasets, which often contain a great deal of repeated content.
Lastly, \citet{zhao2023survey} reviews pretraining data sources, strategies for quality filtering, and the importance of data distribution. 
Their summary corroborates our findings regarding domain composition and quality filtering, in particular.

\vspace{-3mm}
\paragraph{Data, Toxicity, \& Quality}
\label{sec:rw-tox}
Research into the quality and toxicity of datasets and their resulting models has seen mixed findings. 
All of the major models report using significant data pre-processing and toxicity/quality filters, including BERT, T5, BLOOM, OPT, ChinChilla, PaLM, LaMDA, and the GPT-3 series, with the largest of these now using classifiers.
This widespread adoption suggests there are significant implicit benefits, even though they not often externally reported. GLaM does empirically report performance improvements from filtering, particularly on Natural Language Generation (NLG) tasks \citep{du_glam_2021}.

However, in academia, a few works caution against the use of detoxification techniques, including data filters, which can reduce model perplexity on underrepresented communities \citep{xu-etal-2021-detoxifying, welbl2021challenges}.
\citet{welbl2021challenges} also reports that a toxicity classifier reduces toxicity more than than applying data toxicity data filters, but \citet{xu-etal-2021-detoxifying} show this yields the worst perplexity on underrepresented communities.
\citet{meade2022empirical} further corroborates that improvements on bias benchmarks correlates with deteriorations in general language modeling abilities.
Furthermore, investigating GPT-3's described quality filter, \citet{gururangan2022whose} find its quality judgments are unaligned with factuality or literary acclaim but are instead aligned with some notion of langauge ideology more correlated with wealthier zip codes.
Works in the vision domain show data filtering has important detoxification benefits but can reduce performance \citep{nichol2022glide} or introduce other biases \citep{dalle2pretraining2022}.
In summary, pretraining data filters are ubiquitous in the development of non-toxic and high-quality models, but they are prone to reducing their abilities to serve underrepresented communities and may introduce new biases.

Additional work has shown that instruction tuning \citep{chung2022scaling,longpre2023flan} and forms of alignment tuning \citep{ouyang2022training, bai2022constitutional} have both reduced unwanted toxic generation.

\vspace{-3mm}
\paragraph{Data \& Time}
\label{sec:rw-time}
Natural language is known to evolve and change over time \citep{altmann2009beyond,labov2011principles,eisenstein2014diffusion,jaidka-etal-2018-diachronic}.
As language's distribution shifts, the ability of models to perform well on new test sets has also been shown to degrade, due to their static knowledge of recent events, syntactic and semantic practices \citep{lazaridou2021mind,agarwal2022temporal,longpre2021entity}.
\citet{luu2021time,lazaridou2021mind,pmlr-v162-liska22a,yaowild,zhang2021situatedqa,jang2022temporalwiki} offer evaluation sets to measure this phenomena.
Proposed remedies include finetuning on more recent data \citep{luu2021time}, adaptive/continuous pretraining \citep{lazaridou2021mind,rottger-pierrehumbert-2021-temporal-adaptation}, data augmentation \citep{singh2022addressing}, modeling text with its timpestamps \citep{dhingra2022time}.
To our knowledge, no work has thoroughly investigated the effects of temporal degradation when pretraining from scratch.

\vspace{-3mm}
\paragraph{Data \& Domains}
\label{sec:rw-domains}
The composition of public datasets, like C4 and the Pile, is guided mostly by licensing, which severely restricts availability.
Even so, \citet{villalobos2022will,chinchilla2022implications,hoffmann2022training} suggest we are imminently exhausting high-quality text data on the web to train compute-optimal larger LMs, at least with existing training efficiency.
This poses a challenge, given the demonstrated importance of high quality and diverse training data to strong generalization \citep{gao2020pile,papadimitriou2020learning}.
A great deal of literature has dedicated itself to adapting static pretrained models to new downstream domains, using domain adaptive pretraining \citep{gururangan2020don}, finding intermediate finetuning tasks \citep{pruksachatkun2020intermediate}, dynamically balancing data sources \citep{wang2020balancing}, data selection \citep{iter2021complementarity, albalak2023improving}, augmentation \citep{longpre2019exploration}, and active learning \citep{longpre2022active}.
Another line of work demonstrates the potential of pretraining on carefully crafted synthetic data \citep{wuinsights}.

Most similar to this section of our work, \citet{xie2023doremi} re-balance mixtures of the Pile to achieve more performant and efficient convergence. 
\citet{xie2023data} use importance sampling to select subsets of the Pile most useful for target downstream tasks, in lieu of quality filters, to achieve 2\% improvement on downstream tasks.
\citet{pruksachatkun2020intermediate} systematically benchmark the effects of intermediate finetuning tasks, similar to how we benchmark different compositions of pretraining tasks.

\vspace{-3mm}
\paragraph{Model \& Data Scaling}
\label{sec:rw-scaling}
Prior work has explored scaling model size \citep{kaplan2020scaling,tay2022scaling,du_glam_2021}, the amount of pretraining data or the number of pretraining steps \citep{liu2019roberta,chowdhery2022palm,brown2020language}.
Chinchilla investigated and reported optimal compute scaling laws, expressing a relationship between model and data size \citep{chinchilla2022implications}.
Recent work has demonstrated that new abilities emerge at greater scale \citep{wei2022emergent}, but also that many of these benefits can be distilled or compressed into smaller models \citep{alpaca, movva2022combining}.
In this work, we investigate how temporal pretraining misalignment varies on different model sizes, which to our knowledge was previously unanswered.

\vspace{-3mm}
\section{Conclusion}
The relative age of documents, content filters, and data sources each have significant effects on downstream model behaviour.
These effects can be reduced, but not eliminated, by finetuning.
We recommend that model developers and users pay close attention to these details in designing/selecting the model most relevant to their needs, as each decision has a specific, quantifiable trade-off profile. 
For instance, it may be important to decide between improving toxicity identification or reducing toxic generation, performance on brand new or older data sources, and biomedical or books text domains.
These countless choices are inherent in curating any pretraining dataset.
While we are only able to evaluate a small fraction of these, we are able to show which choices matter and by how much, and we hope to inspire further work evaluating dataset composition and predicting behaviors of models given pretraining datasets.
%Our findings future work in the development of open source pretraining corpora collection, in fair evaluating between models scraped at different times, and for the design of new language models.

% \vspace{3mm}
% \noindent

\section*{Acknowledgements}
We would like to thank Daniel Smilkov for his technical assistance in characterizing large corpora, Maarten Bosma and Jacob Andreas for their early guidance on this project, Tom Small for his visual design support, and Noah Constant for feedback on the paper.
This work is supported by NSF \#1652536.

\bibliographystyle{plainnat}
\bibliography{main}

\clearpage
\appendix
\phantomsection
\addcontentsline{toc}{section}{Appendix} 
\part{Appendix} 
\parttoc

\vspace{-3mm}
\section{Contributions}

\begin{itemize}
    \item \textbf{Shayne Longpre} \; Project lead and primary coder. Led experiment design, implementation, pretraining, evaluation, and analysis.
    \item \textbf{Gregory Yauney} \; Core contributor. Led evaluation implementation and analysis on toxicity and quality filtering section (\cref{sec:quality-toxicity}). Contributed code for \cref{sec:data-analysis}. Also supported writing and analysis.
    \item \textbf{Emily Reif} \; Core contributor. Led the analysis of data characteristics pre- and post-curation (\cref{sec:data-analysis}). Also supported writing and analysis.
    \item \textbf{Katherine Lee} \; Core contributor. Supported infrastructure implementation, debugging, overall analysis and framing, especially for \cref{sec:time}.
    \item \textbf{David Mimno} \; Core contributor. A primary advisor, supporting analysis, framing, writing, and particularly discussion of key take-aways and recommendations (\cref{sec:discussion}).
    \item \textbf{Daphne Ippolito} \; Core contributor. The primary advisor and also a coding contributor, supporting both the analysis, framing, and writing as well as running many of the experiments and evaluations.
    \item \textbf{Adam Roberts} \; Supporting advisor, especially on modeling choice and pretraining infrastructure.
    \item \textbf{Barret Zoph} \; Supporting advisor on experiment design.
    \item \textbf{Denny Zhou}\; Supporting advisor on writing and framing.
    \item \textbf{Jason Wei} \; Supporting advisor on experiment design, writing and framing.
    \item \textbf{Kevin Robinson} \; Supporting advisor on toxicity evaluations and their implementation.
\end{itemize}

\vspace{-3mm}
\section{Expanded Literature Review}

Table \ref{tab:model_survey_filter_list} lists popular and well-known models trained in the last several years and a summary of the available information about their training data.

\begin{table}[t]
    \centering
    \small
\caption{Additional notes on each model's filtering details.
}
\label{tab:model_survey_filter_list}
% \begin{tabular}{p{0.5in}|llllllp{0.8in}l|p{2in}ll|l}
\begin{tabular}{l|p{0.8\linewidth}}
\toprule
\textsc{Model} & \textsc{Filtering Details}\\
\midrule
\textsc{Bert} & ``ignore lists, tables, and headers''\\
\textsc{GPT-2} & removed Wikipedia \\
\textsc{RoBerta} & CC filtered to news and Winograd-like subsets\\
\textsc{XLNet} & ``heuristics to aggressively filter out short or low-quality articles''\\
\textsc{T5} & Heuristic quality, toxicity, and length filters; code removed \\
% July 2020 release, 7\% multilingual text according to section 3.3
\textsc{GPT-3} & Filtered based on similarity to high-quality reference corpora. \\
% GPT-J: Aug 2021. Pile: Dec 2020.
\textsc{GPT-J/Neo} & Uses fasttext classifier on Pile-CC, with OpenWebText2 as the high-quality reference. \\
% Dec 2021. Tox filter because they cite T5's C4 directly.
\textsc{GLaM} & Classifier with Wikipedia, books and selected websites as positive examples \\
% Jan 20, 2022.
\textsc{LaMDA} & ``LaMDA SSI and safety discriminators are also used to score and filter 2.5M turns of dialog data sampled from the
pre-training dataset'', which are then trained on.  \\
% Feb 2022 Release. 
% Filtering: (section B.2) and "we
% filtered out all files larger than 1MB or with lines longer than 1000 characters, to exclude automatically
% generated code. We also removed duplicates of the same file, ignoring whitespace in comparisons"
\textsc{AlphaCode} &  Filtering heuristics to exclude automatically generated code \\
% March 2022
\textsc{CodeGen} & Heuristic filters for code quality \\
% march 2022 release
\textsc{Chinchilla} & Heuristic-based quality filtering, SafeSearch filter \\
% June 2022. Filters: Uses 5% PaLM data (same filters), and 95% acad.
\textsc{Minerva} & Same as PaLM for non-academic data \\
% Published Nov 2022. Released July 2022. 29% English data.
% 62% of docs come from ROOTS sec 2: (11% code, 10% academic, 10% books, 5% wiki) , then 38% OSCAR (ML CC)
\textsc{BLOOM} & heuristic-based quality and porn filtering\\
% October 2022 release, Appendix D says data collected up to 2021. 22% multilingual
\textsc{PaLM} & Same as GlaM \\
% Nov 16 2022. Filters: We apply several quality filters, including excluding papers from journals with certain keywords, and also
% excluding papers with a low journal impact factor.
\textsc{Galactica} & Apply several quality filters: exclude papers from journals with certain keywords or low journal impact factor \\
% May 2022, appendix C.2 doesn't give enough info for percentages
% \textsc{OPT} &  &  &  &  &  & & Pile & \ &  &  &  & ? \\
% Feb 2023 release. data up to 2020. Used multilingual wikipedia (~4% all data)
\textsc{LLaMA} & Classifier to filter out low-quality and un-Wikipedia-like text \\
\bottomrule
\end{tabular}
\end{table}

% LaMDA&          No&             Yes&        Yes&        Yes&    No&     Yes&            n/a&            No& none mentioned & no & TODO \\
% Gopher/Chinchilla&No&           Yes&        No&         Yes&    Yes&    No&             news&           No& Heuristic-based quality filtering, SafeSearch filter & No & TODO \\
% GLaM&           No&             Yes&        Yes&        No&     Yes&    Yes&            n/a&            No& Classifier with Wikipedia, books and selected websites as positive examples & No & TODO \\
% PaLM&           No&             Yes&        Yes&        Yes&    Yes&    Yes&            news&           Yes&Same as GlaM & No & TODO \\
% GPT-3&          Yes&            Yes&        No&         No&     Yes&    No&             n/a&            No&Filtered based on similarity to high-quality reference corpora & No & TODO \\

% OPT&            Yes&            Yes&        Yes&        Yes&    Yes&    Yes&            same as GPT-Neo&No& Same as RoBERTa & Partially & TODO\\
% LLAMA&          Yes&            Yes&        Yes&        Yes&    Yes&    Yes&            academic& Yes& Classifier to filter out low-quality and un-Wikipedia-like text &Partially&TODO\\
% BLOOM&          TODO&           TODO&       TODO&       TODO&   TODO&   TODO&           TODO&   Yes&   heuristic-based quality and porn filtering& Yes& TODO\\
% \bottomrule
% \end{tabular}
% \end{table}

% \end{landscape}

\vspace{-3mm}
\section{Experimental Details}
\label{sec:app-exp-details}
This section provides further details on the methodology and hyperparameter settings used for pretraining, finetuning, and evaluation.

To allow for a model that can generate without finetuning but also perform well after finetuning, we rely on the extensive experiments of \citet{wang2022language}.
Their empirical results suggest these criteria are met with a Causal Decoding architecture with a Full Language Modeling pretraining objective (``CD-FLM''), which permits generation without finetuning, followed by a Prefix Language Modeling objective (PLM) for finetuning, where the causal attention mask is removed from the original prompt.

\vspace{-3mm}
\subsection{Pretraining Details}
\label{app:pretrain-hps}

Our two pretraining datasets are C4 \citep{raffel2020exploring} and the Pile \citep{gao-etal-2021-making}.
We use the same vocabulary for both as used in the original T5 from \citet{raffel2020exploring}.
All training is conducted using T5X \citep{roberts2022t5x} and Tensorflow \citep{abadi2016tensorflow} on TPUs.
Specific hyperparameters for \bigLM{} and \smalLM{} pretraining are detailed in \cref{tab:pretrain-hyperparams}.

\begingroup
\setlength{\tabcolsep}{4pt}

\begin{table*}[ht]
    \centering
    \small
    % \begin{tabular}{>{\raggedright}p{7cm} crll}
    \caption{
    \textbf{Pretraining hyperparameters} We adopt default pretraining hyperparameters from \citet{wang2022language}, who select their parameters to fairly compare across a wide range of T5-based pretraining and architecture experiments.}
    \begin{tabular}{l | r | r}
    \toprule
    \textsc{Parameter} & \textsc{\bigLM} & \textsc{\smalLM} \\
    \midrule
    TPUs & 8x8x8 & 8x8 \\
    Batch Size & 4096 & 4096 \\
    Sequence Length & 512 & 512 \\
    Training Steps & 88,064 & 88,064 \\
    Dropout & 0.0 & 0.0 \\
    \midrule
    Base Learning Rate & \multicolumn{2}{c}{0.5} \\
    Decay Factor & \multicolumn{2}{c}{0.5} \\
    Warmup Steps & \multicolumn{2}{c}{1000} \\
    Steps per Decay & \multicolumn{2}{c}{20000} \\
    \bottomrule
    \end{tabular}
    \label{tab:pretrain-hyperparams}
\end{table*}
\endgroup

\vspace{-3mm}
\subsection{Finetuning Details}
\label{app:finetune-toxicity-hps}
Unless otherwise noted, evaluation was performed by finetuning on the train set for each benchmark task, and then evaluating on either the validation or test set (specified in each section).
Finetuning hyperparameters are given in \cref{tab:finetune-tox-identification-hyperparams}.

\begingroup
\setlength{\tabcolsep}{4pt}
\begin{table*}[ht]
    \centering
    \small
    % \begin{tabular}{>{\raggedright}p{7cm} crll}
    \caption{
    \textbf{Finetuning and Evaluation Parameters for each set of Downstream Tasks.} We report the finetuning hyperparameter settings and evaluation metric used for finetunting and evaluating the pretrained models. We conduct finetuning for four sets of tasks: toxicity identification tasks (Toxigen, Social Bias Frames, and DynaHate), Natural Questions (for pretraining domain transfer analysis), general NLU performance (SuperGLUE), and the Time tasks (including PubCLS, NewSum, PoliAff, TwiERC, and AIC). For T5 Small models, we modify the number of training steps accordingly, as shown in the last row.}
    \begin{tabular}{l | cccc}
    \toprule
    \textsc{Parameter} & \textsc{Tox-Identify} & \textsc{Natural Qs} & \textsc{SuperGLUE} & \textsc{Time} \\
    \midrule
    \multicolumn{5}{c}{\textsc{\bigLM}} \\
    \midrule
    TPUs & 8x8 & 8x8 & 8x8 & 8x8 \\
    Sequence Length & 128 & 512 & 512 & 128 \\
    Batch Size & 128 & 128 & 128 & 128 \\
    Dropout & 0.1 & 0.1 & 0.1 & 0.1 \\
    Training Steps & 10k & 50k & 100k & See \cref{tab:time-hyperparams} \\
    Learning Rate & 1e-3 & 1e-3 & 1e-3 & See \cref{tab:time-hyperparams} \\
    Eval Metric & AUC-ROC & Acc & (By Dataset) & See \cref{tab:time-hyperparams} \\
    \midrule
    \multicolumn{5}{c}{\textsc{\smalLM} \emph{(where different)}} \\
    \midrule
    Training Steps & 30k & 50k & 100k & See \cref{tab:time-hyperparams} \\
    \bottomrule
    \end{tabular}
    \label{tab:finetune-tox-identification-hyperparams}
\end{table*}
\endgroup

\vspace{-3mm}
\subsection{Toxicity Evaluation Details}
\label{app:toxicity-details}

In this section we describe the evaluation details for the results presented in \cref{sec:quality-toxicity}.

\vspace{-3mm}
\paragraph{Toxicity Identification}
For toxicity identification evaluation, we finetune separately on each of the following datasets: Social Bias Frames (SBF, \citealp{sap-etal-2020-social}), DynaHate (DH, \citealp{vidgen-etal-2021-learning}), and Toxigen \citep{hartvigsen2022toxigen}.
We then evaluate on their respective test sets, using AUC-ROC, which computes the error rates over different classification thresholds.
Note that each of these datasets has a slightly different version of toxicity they are evaluating for: biases and offensive stereotypes implied in language for Social Bias Frames, types of hateful speech like animosity, dehumanization, and derogation for DynaHate, and adversarial, subtle and implicit forms of toxicity towards minority groups (e.g., stereotyping, microaggressions), for Toxigen.
We average over them to understand a model's capabilities in identifying toxicity broadly.

\vspace{-3mm}
\paragraph{Toxic Generation}
For toxic generation, we do not do any finetuning.
Instead, we evaluate how often each model generates 
toxic continuations.
To measure a model's propensity to generate sexual, profane, or toxic responses to a mix of benign and malicious inputs, we use prompts from the RealToxicityPrompts \citep{gehman2020realtoxicityprompts} dataset, as well as the prompt set developed to explore representational bias in PaLM \citep{chowdhery2022palm}.

For each model, we generate 25 responses per prompt, using top-$k$ sampling ($k=40$), with a temperature of $1.0$.
Following \citet{chowdhery2022palm}, we use the Perspective API to score responses, labeling those with a toxicity score $>=0.5$ as toxic.
We compute the probability one of these generations per prompt is toxic, for two evaluation splits, one with non-toxic inputs, and one with toxic inputs.
The resulting evaluation metrics are denoted RPT-T (for Toxic inputs) and RPT-NT (for Non-Toxic inputs).

The Representational Bias benchmark was constructed from the identity terms in \citep{rae2021scaling} and is described in more detail in \citet{chowdhery2022palm}.
It uses templates constructed for several identity markers, for which we use the subset related to gender, race, and religion.
Following the practice with RealToxicityPrompts, we sample 50 responses per prompt, use top-$k$ sampling ($k=40, \text{temperature}=1.0$), and then classify a response as toxic if Perspective API evaluates its toxicity score $>=0.5$.
We average the toxicity rates per identity marker and per example to compute the overall RepBias score, where higher indicates more toxic responses were produced on average.
We also compute the 95\% confidence interval to show where changes in mean are significant.

\vspace{-3mm}
\subsection{Time Evaluation Details}
\label{app:time-eval-details}
\label{app:temp-deg-measure}

This section describes the evaluation details for the results presented in \cref{sec:time}.
In applied settings, the available training data (either for pretraining or finetuning) may be from different years than the test-time data.
To mimic these situations, \citet{luu2021time} construct several datasets segmented by the year they are collected from in order to measure the performance impact of differences in the time of collection of finetuning and evaluation splits.
As described in \cref{sec:time-eval}, we select 5 of the datasets that are shown to be quite sensitive to these temporal misalignments, and that cover different tasks and data sources.
These tasks are summarization, named entity recognition, classifying political affiliation, classifying academic topic, and classifying the news source.

Due to the unique nature of each of these tasks in the temporal degradation experiments, we simply finetune on each task individually, before evaluating on their respective test sets.
For each dataset, we finetune using 4x4 TPUs with a batch size of $64$, a maximum sequence length of $128$, and we validate every 500 training steps.
We select the test set score with the highest validation accuracy across training.
The best learning rate and the total number of steps required to reach convergence varied by model and model size, and are reported in \cref{tab:time-hyperparams}.
These hyperparameters are chosen based on initial experiments attempting to produce stable learning curves which peak near the values observed in \citet{luu2021time}.

\begingroup
\setlength{\tabcolsep}{4pt}
\begin{table*}[ht]
    \centering
    \small
    % \begin{tabular}{>{\raggedright}p{7cm} crll}
    \caption{
    \textbf{Time Dataset \& Training Details}: For each of the five datasets used to evaluate the model's ability over different temporal periods, we report the learning rate and number of steps used in each model size. These hyperparameters were chosen to ensure consistent convergence and stability within our infrastructure settings.}
    \begin{tabular}{l | ll | rr | rr}
    \toprule
    & & & \multicolumn{2}{c}{\textsc{\bigLM}} & \multicolumn{2}{c}{\textsc{\smalLM}} \\
    \midrule
    \textsc{Domain} & \textsc{Task} & \textsc{Metric} & \textsc{LR} & \textsc{Steps} & \textsc{LR} & \textsc{Steps} \\
    \midrule

    \multirow{2}{*}{\textsc{News}} & \textsc{PubCLS} & Acc & 1e-4 & 30k & 1e-3 & 30k \\
    & \textsc{NewSum} & Rouge-L & 5e-4 & 40k & 1e-3 & 40k \\
    \midrule
    \multirow{2}{*}{\textsc{Twitter}} & \textsc{PoliAff} & Acc & 1e-4 & 15k & 1e-4 & 15k \\
    & \textsc{TwiERC} & Acc & 1e-4 & 30k & 1e-3 & 30k \\
    \midrule
    \textsc{Science} & \textsc{AIC} & Acc & 1e-4 & 30k & 1e-3 & 60k \\
    \bottomrule
    \end{tabular}
    \label{tab:time-hyperparams}
\end{table*}
\endgroup

We follow \citet{luu2021time}'s exact prescription in calculating Temporal Degradation (TD), as well as their reported Pearson correlation measurements ($r$).
Temporal degradation can be interpreted as the average rate of deterioration in performance for a time period, measured in years.
Since a temporal deterioration score is calculated per evaluation year, we average over all evaluation years to compute a final TD score for a dataset.
Furthermore, each dataset has a different span of available training and evaluation years.
To account for this, we follow \citet{luu2021time} in presenting the Pearson correlation coefficient, which presents the strenght of the relationship between time differences and performance deterioration.
We also replicate the Wald test with null hypothesis that the slope is zero.

For evaluating the temporal degradation of pretraining, TD$_p$, we modify \citet{luu2021time}'s original formula to measure the different $D(t'\rightarrow t)$ where $t'$ is now the pretraining year.
However, in this setting, performance samples are represented with different finetuning years. 
To account for this, we only compare the relative performance changes of the pretraining year $t_p$, against models with the same finetuning $t_f$ and evaluation years $t_e$.
In other words, given $S_{t_p \rightarrow t_f \rightarrow t_e}$, we will only compare its performance to $S_{t_p' \rightarrow t^f \rightarrow t^e}$ where $t_p' \ne t_p$, but $t^f$ and $t^e$ are fixed to their respective values.

$$D(t_p' \rightarrow t_e) = - (S_{t_p' \rightarrow t_f'' \rightarrow t_e} - S_{t_p \rightarrow t_f'' \rightarrow t_e}) * \text{sign}(t_p' - t_e)$$

In some edge cases, there is no evaluation year equivalent to a pretraining year, $\forall t \in T, t_p \ne t_e$, and so the term $S_{t_p \rightarrow t_f'' \rightarrow t_e})$ does not exist.
In this case, we set this term to be the one where $t_p$ and $t_e$ are closest.
And, as before, the precise term used will depend on which version of $t_f$ is being calculated for.

\vspace{-3mm}
\subsection{Evaluating Domains with Question Answering Datasets}
\label{app:domain-eval-details}

This section describes the evaluation details for the results presented in \cref{sec:domain-comp}.
These experiments involve pretraining models with different subsets of the corpora from the Pile \citep{gao2020pile} and seeing the effects on a variety of downstream evaluation domains, represented by question answering datasets.
As such, we are able to map the effects of pretraining domains to evaluation domains.

First, we discuss the construction of the pretraining domains.
We partition the Pile's source datasets into categories representing thematically similar sources of data, as seen in \cref{tab:pile-partitions}.
We refer to these categories as Domains.
These domain partitions are subjective and cannot perfectly separate out text into these categories. 
For instance, Wikipedia, Books, and Common Crawl data inevitably contain some Academic information, but overall these partitions represent distinct features (see \cref{sec:data-analysis}) that we have attempted to delineate by areas of interest to practitioners and researchers.
Prior work has attempted to measure, emphasize, or target (either for inclusion or exclusion) the particular categories of data we've used in our partitions, such as more books and structured data \citep{brown2020language, chowdhery2022palm}, code data \citep{chen2021evaluating}, and legal data \citep{dodge2021documenting}, among others.
% For instance, Wikipedia has served a unique and prominent role in the NLP's training and evaluation norms, common crawl represents a permissively licensed set of web pages \citep{raffel2020exploring}, OpenWeb    have suggested there are particular benefits to training on Books data, and other work has focused in on 

\begingroup
\setlength{\tabcolsep}{4pt}
\begin{table*}[ht]
    \centering
    \small
    % \begin{tabular}{>{\raggedright}p{7cm} crll}
    \caption{
    \textbf{Partitions of the Pile's Data Sources into Domains} The Pile contains 22 distinct sources of data, which we manually partition into 9 thematically similar domain clusters.}
    \begin{tabular}{l | p{0.3\linewidth} r p{0.45\linewidth}}
    \toprule
    \textsc{Category} & \textsc{Components} & \textsc{Size} & \textsc{Description} \\
    \midrule
    \textsc{CC} & Pile-CC & 227 GB & A filtered set of Common Crawl websites, scraped with JusText \citep{endredy2013more}. \\
    \textsc{OpenWeb} & OpenWebText2 & 63GB & Scraped OpenWebTextCorpus using upvoted Reddit outgoing links.\\
    \textsc{Wikipedia} & Wikipedia (en) & 6 GB & The English scrape of Wikipedia. \\
    \textsc{Books} & Books3, BookCorpus2, Gutenberg (PG-19) & 118 GB & The Bibliotik general literature collection, PG-19's pre-1919 western classics, and BookCorpus's set of yet unpublished works. \\
    \textsc{PubMed} & PubMed Central, PubMed Abstracts & 109 GB & Biomedical articles from 1946 to present \\
    \textsc{Academic} & ArXiv, PhilPapers, NIH ExPorter & 60 GB &  Preprint academic papers in Math, Computer Science, Physics, and Philosophy.\\
    \textsc{Code \& Math} & Github, StackExchange, DM Mathematics & 135 GB & Code repositories, documentation, coding questions and answers, and mathematical problems. \\
    \textsc{Legal} & FreeLaw, USPTO Backgrounds & 74 GB & Court filings, judicial opinions, and patents \\
    \textsc{Social} & Ubuntu IRC, EuroParl, Enron Emails, HackerNews, OpenSubtitles, YoutubeSubtitles & 33 GB & Movie and video subtitles, chat logs, emails, and text from social news websites. \\
    \midrule
    \textsc{Base} & All & 825 GB & A wide mix of online text from the web, wikipedia, books, academic articles, code, legal, and social sources. \\
    \bottomrule
    \end{tabular}
    \label{tab:pile-partitions}
\end{table*}
\endgroup

The Domains of the Pile were then each separately ablated from pretraining to understand the effect of their absence.
To evaluate their absence on the performance of downstream domains, we chose to use the question answering task expressly because there is a wide variety of similarly formatted evaluation datasets available.
For these question answering datasets we train only on Natural Questions \citep{kwiatkowski2019natural}, a popular QA dataset, to teach the model the general task.
For evaluation, as described in \cref{sec:finetune-eval}, we use UnifiedQA \citep{khashabi-etal-2020-unifiedqa} and MRQA \citep{fisch2019mrqa}'s collection of datasets to evaluate how each pretrained model performs on a given ``domain'', or set of datasets with similar source characteristics.
We partition the question answering datasets from UnifiedQA and MRQA into five categories.
Datasets with Wikipedia documents represented in their collection are assigned to the \textsc{Wiki} category, datasets with scraped web documents or news are assigned to the \textsc{Web} category, and so on.
Datasets may belong to multiple categories, depending on how they were constructed.
The question answering evaluation partitions are shown in \cref{tab:qa-partitions}.
Finally, we evaluate on each question answering dataset and report the average F1 score for each category.

\begingroup
\setlength{\tabcolsep}{4pt}
\begin{table*}[ht]
    \centering
    \small
    % \begin{tabular}{>{\raggedright}p{7cm} crll}
    \caption{
    \textbf{Partitions of Question Answering evaluation datasets from the UnifiedQA \citep{khashabi-etal-2020-unifiedqa} and MRQA \citep{fisch2019mrqa} collections.} To evaluate the performance of pretraining strategies on different text domains, we assign datasets into categories corresponding to their source material:web-based, wikipedia, academic, biomedical, or and/books). Certain datasets are also designed specifically to test advanced common sense reasoning, or decision boundaries using contrast sets \citep{gardner2020evaluating}. Datasets can belong to multiple categories.}
    \begin{tabular}{l | p{0.4\linewidth} p{0.4\linewidth}}
    \toprule
    \textsc{Category} & \textsc{Datasets} & \textsc{Description} \\
    \midrule
    \textsc{Wiki} & AmbigQA, DROP, HotpotQA, NaturalQuestions, Quoref, RelationExtraction, ROPES, SearchQA, SQuAD-1, SQuAD-2, TriviaQA  & Datasets with Wikipedia text. \\
    \textsc{Web} & AmbigQA, CommonsenseQA, DuoRC, NaturalQuestions, NewsQA, SearchQA, TriviaQA  & Datasets partially sourced or collected from the web, including user logs and news.  \\
    \textsc{Books} & NarrativeQA  & A dataset sourced from books. \\
    \textsc{BioMed} & ARC-Easy, ARC-Hard, BioASQ, TextbookQA  & Datasets with high-school or graduate level scientific or medical content. \\
    \textsc{Academic} & AI2-Elementary-Science, ARC-Easy, ARC-Hard, RACE, ROPES, TextbookQA & General academic data and exams. \\
    % \textsc{Social} & TweetQA & Data collected from social media, user generated text. \\
    \midrule
    \textsc{Common Sense} & CommonsenseQA, PhysicalIQA, SocialIQA & Datasets which test common sense reasoning. \\
    \textsc{Contrast Sets} & Contrast-Set-DROP, Contrast-Set-Quoref, Contrast-Set-ROPES & Datasets re-configured as Contrast Sets \citep{gardner2020evaluating}, which are manual perturbations to make examples more challenging.\\
    \bottomrule
    \end{tabular}
    \label{tab:qa-partitions}
\end{table*}
\endgroup

% \subsection{General Natual Language Understanding Evaluation Details}
% \label{app:superglue-eval}

% To evaluate a model's performance on traditional NLP tasks more generally, we use SuperGLUE \citep{wang2019superglue}.
% We chose to train on the training splits for 8 SuperGLUE tasks combined, before evaluating on each of the 8 tasks separately.
% These tasks are BoolQ (accuracy), CB (accuracy), CoPA (accuracy), MultiRC (F1), ReCoRD (F1), MultiRC (F1), RTE (accuracy), WiC (accuracy), WSC (accuracy).

\clearpage
\vspace{-3mm}
\section{Impact of Data Curation on Data Composition: Further Analysis}
\label{sec:app-data-analysis}

\vspace{-3mm}
\subsection{Feature Definitions}

As discussed in \cref{sec:data-analysis}, we calculated a set of features across all datapoints to better understand the distribution shifts for each ablation. The full list of features is as follows:
\begin{itemize}
    \item \textbf{Profanity, Toxicity, and Sexually Explicit} The Perspective API classifies text as violating or passing each of these categories, as described in \cref{sec:data-filters}.
    \item \textbf{Text Quality} The same bag-of-words-based linear classifier as used in PaLM \citep{chowdhery2022palm} and GLaM \citep{du_glam_2021}, is used to distinguish between text that looks like Wikipedia and books from other text, as described in \cref{sec:data-filters}.
    \item \textbf{Personally Identifiable Information (PII)} A basic classifier, similar to \citet{gcloud_info}, detects the presence of four categories of personally identifiable information: \textbf{names}, \textbf{phone numbers}, \textbf{addresses}, and \textbf{emails}.
    \item \textbf{Readability} The Flesch–Kincaid readability test \citep{kincaid1975derivation} is applied to each document, assigning documents a grade level based on the number of words per sentence and number of syllables per word.
    \item \textbf{Average Word Length} Measured in characters.
    \item \textbf{Document Length} Measured in characters.
    \item  \textbf{Non-ASCII Characters} Measured as a percentage of all characters in the document.
    \item \textbf{All-caps Words} Measured as a percentage of all words in the document.
    \item \textbf{Type-Token Ratio} A measure of the lexical diversity, or the ratio of unique tokens to total tokens \citep{bender2013linguistic}.
    \item \textbf{Sentiment} The score assigned by a classifier similar to \citet{gcloud_sentiment}, evaluating the overall sentiment of the text along a spectrum from positive to negative.
\end{itemize}

\begin{figure}[ht]
\centering
\qquad\subfigure[Time in C4]{
\includegraphics[height=13cm]{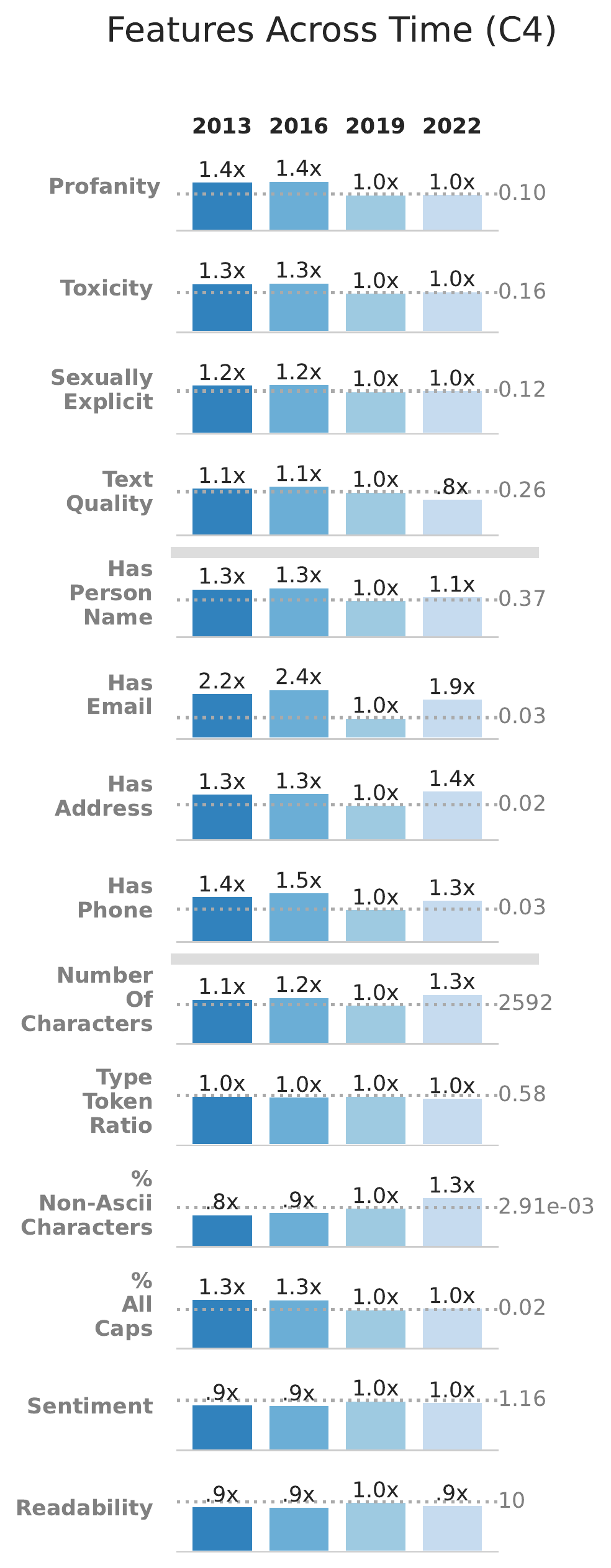}}
\subfigure[C4 vs the Pile]{
\includegraphics[height=13cm]{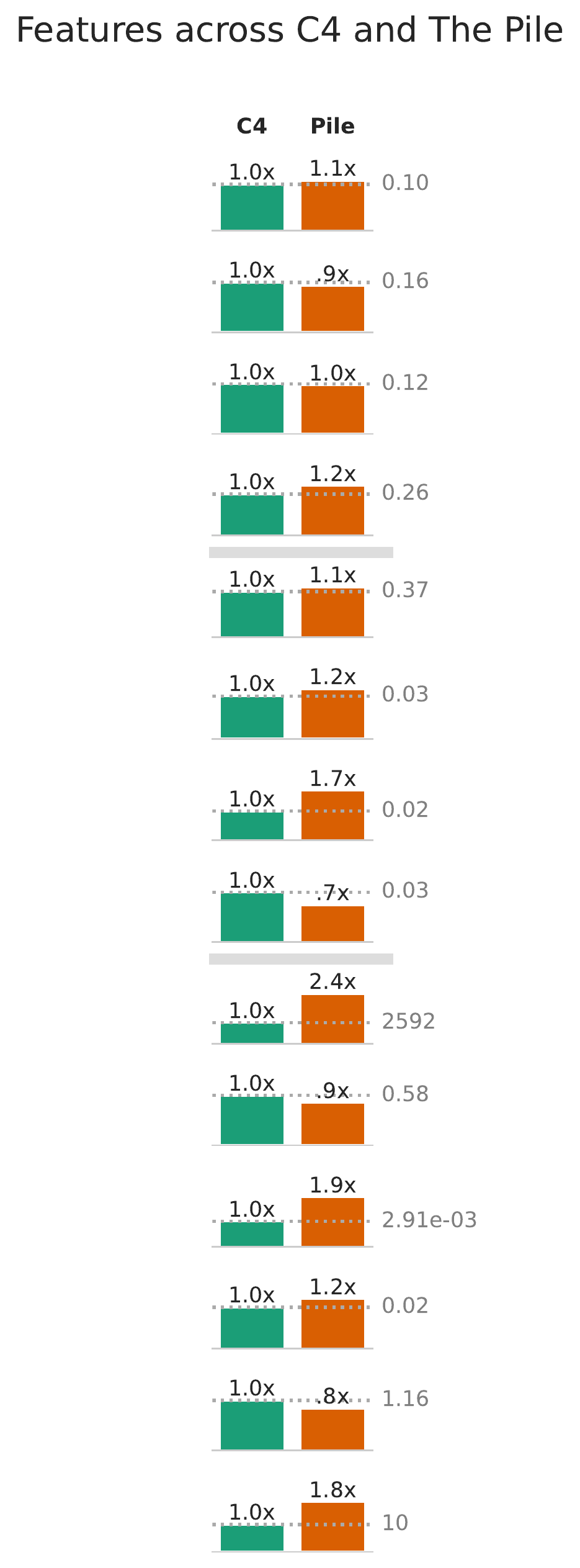}}
    \caption{
    \small \textbf{Feature differences across C4 and the Pile, and time snapshots of C4} Bar height indicates average feature value of each dataset, except for the PII categories which show the fraction of datapoints containing that PII type. The numbers are the fraction difference between the dataset and the baseline, which in this case is C4. The gray dashed line and gray number show the actual value for the baseline.}
        \label{fig:c4-pile-composition}
        \vspace{-3mm}
\end{figure}

\begin{figure}[ht]
\begin{minipage}{0.48\columnwidth}
    \centering
    \includegraphics[width=\columnwidth]{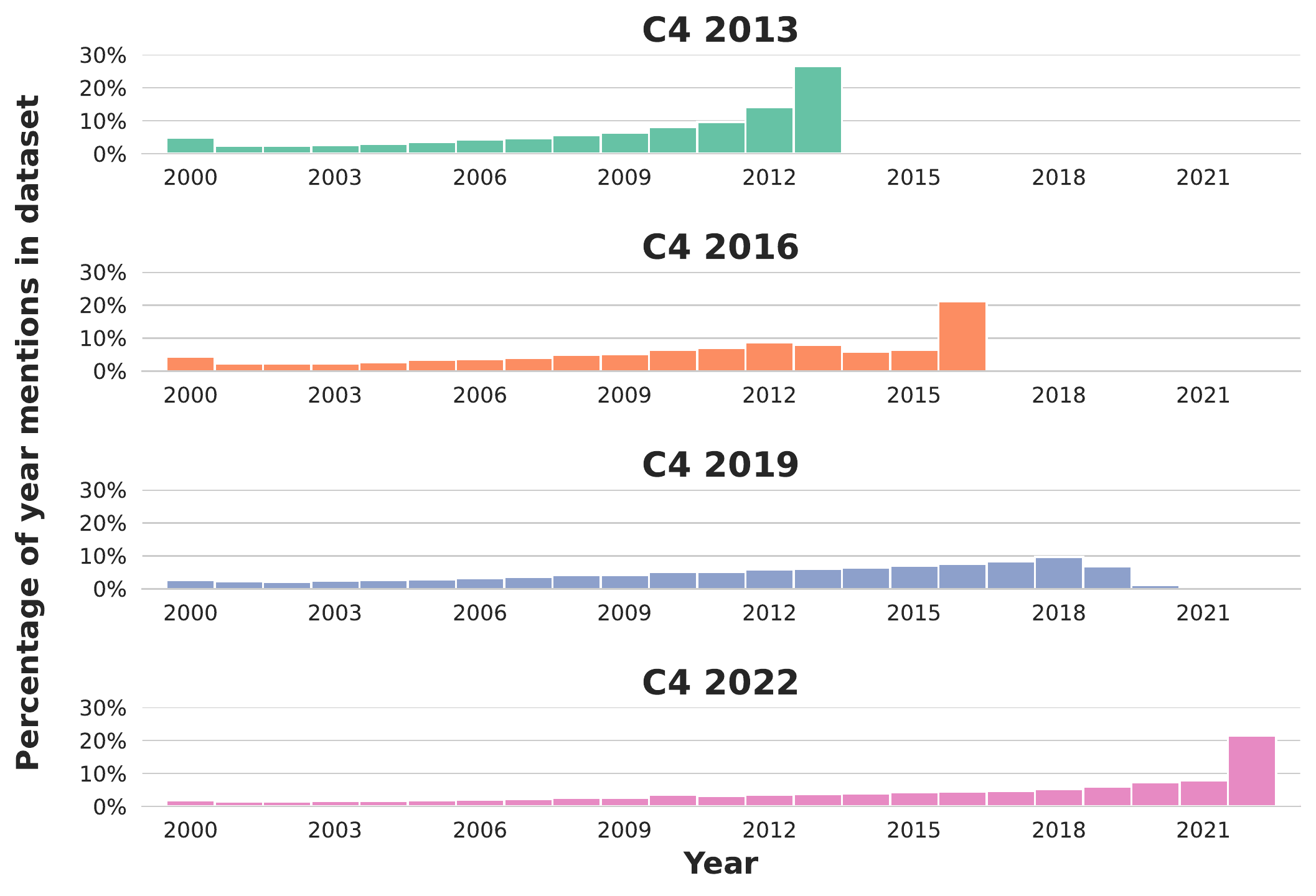}
    \caption{
    \small
    \textbf{Date instances in each of the C4 temporal pretraining versions.}}
    \vspace{-3mm}
    \label{fig:dates-histogram}
\end{minipage}
\hfill
\begin{minipage}{0.48\columnwidth}
    \vskip 1.4cm
    \centering
    \includegraphics[width=\columnwidth]{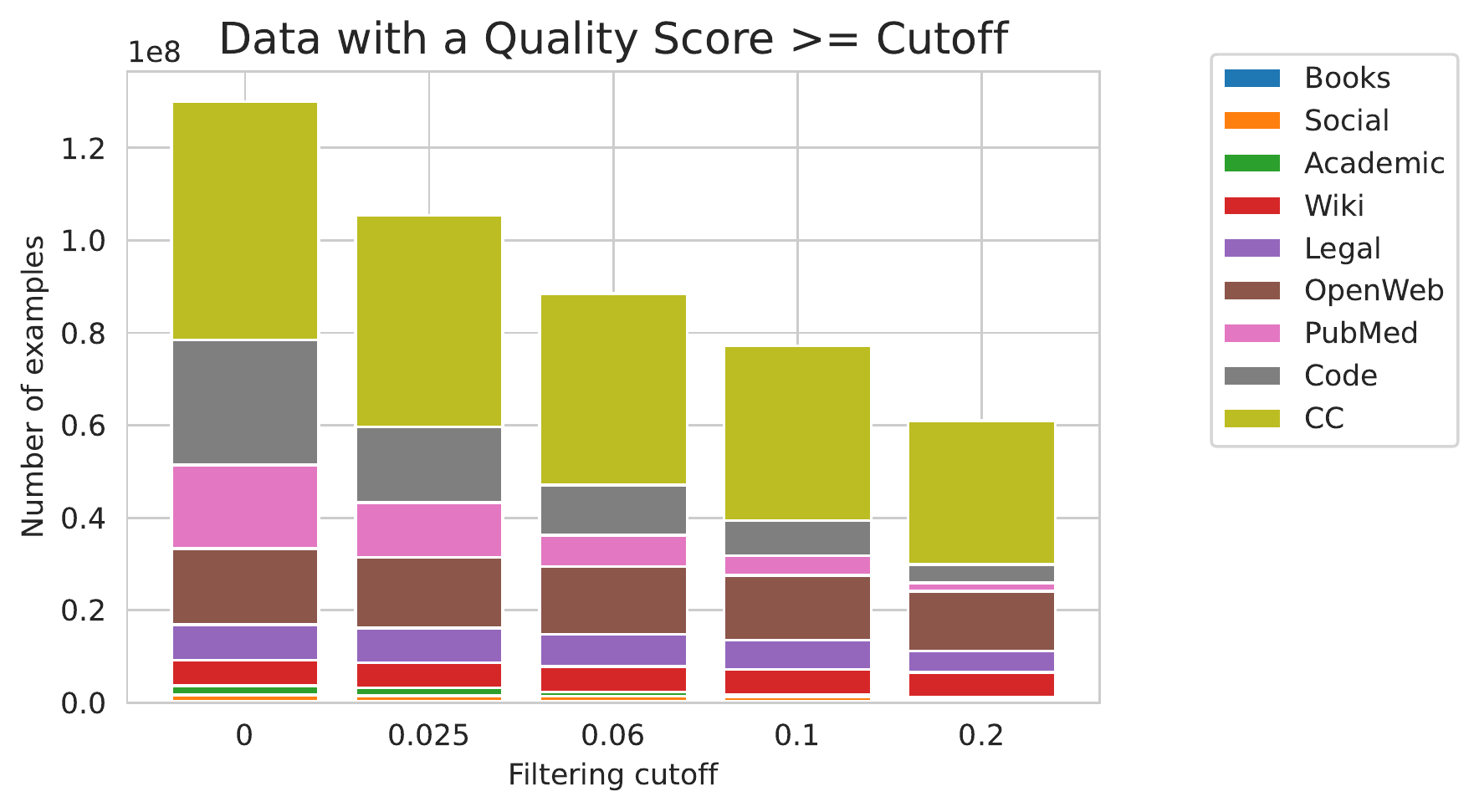}
    \caption{
    \small
    \textbf{Breakdown of domains in the Pile after filtering for multiple quality cutoffs.}\\
    }
    \vspace{-3mm}
    \label{fig:domains_quality_counts}
\end{minipage}
\end{figure}

\vspace{-3mm}
\paragraph {Temporal information in pretraining data} 
While we collected versions of C4 at four different years, each of these versions may also contain data from prior years. 
We estimate the temporal information in the pretraining data by counting instances of dates from 2000 to 2025 in each corpus. 
We do see that there are many mentions of the year of collection, with a quick dropoff of about 5 years earlier (see \cref{fig:dates-histogram}). 
This is necessarily a limited experiment as an article written in 2016 may still mention something occurring in the future in 2019. 
However, since website creation dates are not part of the web-scrape, we use this as a proxy to estimate website creation dates. 

\vspace{-3mm}
\subsection{Breakdown of the Quality Filter on Pile Domains}

While the quality filters are typically applied to large, heterogeneous datasets such as C4, we also ran the quality classifier on the Pile to get a better understanding of what types of datapoints actually passed the quality filtering thresholds. The results are shown in \cref{fig:domains_quality_counts}.

\clearpage
\vspace{-3mm}
\section{Experimental Results}
\label{sec:app-exp-results}

In this section, we lay out the raw results for our toxicity, quality, and temporal degradation evaluations, spanning several evaluation datasets.

\vspace{-3mm}
\subsection{Temporal Degradation Results}
\label{app:time-results}

\begin{figure}[ht]
    \centering
    \includegraphics[width=0.99\linewidth]{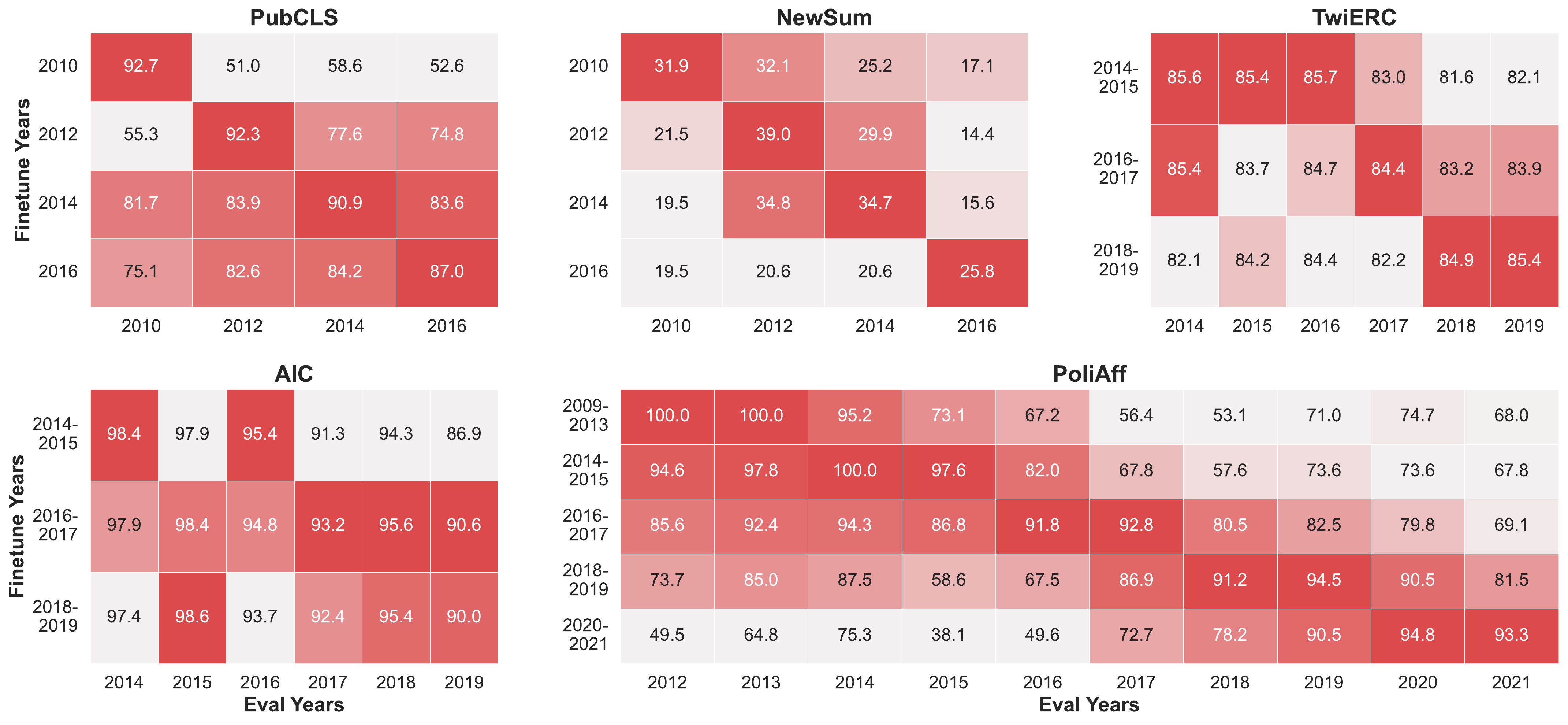}
    \caption{
    \small
    \textbf{A replication of how temporal misalignment in finetuning affects task performance \citep{luu2021time}. In contrast to \cref{fig:time-heatmap-xl-pt}, which shows the effects of pretraining misalignment, this figure focuses on the more well established effect of finetuning misalignment.}\\
    % \textsuperscript{\textdagger} indicates concurrent work.\\
    }
    \vspace{-3mm}
    \label{fig:time-heatmap-xl-ft}
\end{figure}

\citet{luu2021time} measure the temporal degradation due to finetuning and evaluation misalignment.
Before attempting to evaluate misalignment effects specifically for \emph{pretraining}, we mimic their finetuning experiments.
\cref{fig:time-heatmap-xl-ft} shows our results, which corroborate the findings of \citep{luu2021time}.

Next we share the original evaluation results from which we computed the temporal degradation values for both finetuning and pretraining.
These contain a cross-section of the scores produced using a given pretraining year ($y$-axis), finetuning year(s) ($y$-axis), for an evaluation year ($x$-axis).
These results, \cref{tab:time_aic,tab:time_pubcls,tab:time_poliaff,tab:time_twierc,tab:time_newsum}, are provided for both \bigLM{} and \smalLM, for comparison.

\begin{table}[!htb]
    % \caption{Global caption}
    \caption{
    \emph{Left:} Full results on the \textbf{PubCLS} temporal task splits from \citep{luu2021time}. This task evaluates news article source classification, measured with Accuracy. \emph{Right:} Full results on the \textbf{NewSum} summarization task temporal splits from \citep{luu2021time}, evaluated in Rouge-L.}
    \begin{minipage}[t]{\dimexpr.5\textwidth+1em}
    % \begin{minipage}{.4\linewidth}
    %   \caption{}
      \centering
      \small
      \begin{tabular}{ll | cccc}
      \toprule
    \textsc{Pretrain} & \textsc{Finetune} & \multicolumn{4}{c}{\textsc{Eval Time}} \\
    \textsc{Time} & \textsc{Time} & \textsc{2010} & \textsc{2012} & \textsc{2014} & \textsc{2016} \\

    \midrule
    \multicolumn{6}{c}{\textsc{\bigLM}} \\
    \midrule
    
    \multirow{4}{*}{\textsc{2013}} & 2010 & 93.7 & 51.9 & 58.4 & 52.5 \\
    & 2012 & 60.2 & 94.6 & 78.4 & 75.6 \\
    & 2014 & 83.1 & 85.6 & 90.8 & 84.8 \\
    & 2016 & 78.7 & 84.7 & 86.2 & 87.6 \\
    \hline
    \multirow{4}{*}{\textsc{2016}} & 2010 & 93.8 & 51.6 & 59.2 & 53.2 \\
    & 2012 & 55.2 & 93.9 & 79.5 & 77.0 \\
    & 2014 & 81.5 & 86.2 & 92.8 & 85.6 \\
    & 2016 & 76.9 & 82.9 & 84.3 & 89.6 \\
    \hline
    \multirow{4}{*}{\textsc{2019}} & 2010 & 92.9 & 50.6 & 58.6 & 52.2 \\
    & 2012 & 53.4 & 90.5 & 75.9 & 72.9 \\
    & 2014 & 81.3 & 83.2 & 90.6 & 82.8 \\
    & 2016 & 72.3 & 81.1 & 83.4 & 84.8 \\
    \hline
    \multirow{4}{*}{\textsc{2022}} & 2010 & 90.5 & 49.9 & 58.4 & 52.4 \\
    & 2012 & 52.4 & 90.4 & 76.4 & 73.9 \\
    & 2014 & 80.9 & 80.7 & 89.3 & 81.1 \\
    & 2016 & 72.3 & 81.7 & 83.0 & 86.1 \\
    
    \midrule
    \multicolumn{6}{c}{\textsc{\smalLM}} \\
    \midrule
    
    \multirow{5}{*}{\textsc{2013}} & 2010 & 92.9 & 51.9 & 60.2 & 54.1 \\
    & 2012 & 55.4 & 93.3 & 75.7 & 75.9 \\
    & 2014 & 78.2 & 81.9 & 89.9 & 82.5 \\
    & 2016 & 70.5 & 80.0 & 80.7 & 87.4 \\
    \hline
    \multirow{5}{*}{\textsc{2016}} & 2010 & 93.0 & 51.8 & 58.8 & 53.2 \\
    & 2012 & 56.7 & 92.9 & 77.7 & 75.5 \\
    & 2014 & 77.3 & 80.2 & 89.6 & 81.4 \\
    & 2016 & 69.9 & 80.1 & 82.1 & 87.7 \\
    \hline
    \multirow{5}{*}{\textsc{2019}} & 2010 & 92.9 & 51.3 & 59.2 & 53.0 \\
    & 2012 & 58.9 & 93.3 & 76.4 & 75.6 \\
    & 2014 & 78.4 & 82.1 & 90.2 & 82.7 \\
    & 2016 & 69.8 & 81.4 & 80.8 & 87.7 \\
    \hline
    \multirow{5}{*}{\textsc{2022}} & 2010 & 93.3 & 51.6 & 59.1 & 53.2 \\
    & 2012 & 56.2 & 93.2 & 75.6 & 75.1 \\
    & 2014 & 76.4 & 81.0 & 90.1 & 81.7 \\
    & 2016 & 67.8 & 80.4 & 80.1 & 86.8 \\

    \bottomrule
    \end{tabular}
    % \caption{
    \label{tab:time_pubcls}
      
    \end{minipage}%
    % \begin{minipage}{.6\linewidth}
    \begin{minipage}[t]{\dimexpr.5\textwidth+1em}
      \centering
      \small
    \begin{tabular}{ll | cccc}
    \toprule
    \textsc{Pretrain} & \textsc{Finetune} & \multicolumn{4}{c}{\textsc{Eval Time}} \\
    \textsc{Time} & \textsc{Time} & \textsc{2010} & \textsc{2012} & \textsc{2014} & \textsc{2016} \\

    \midrule
    \multicolumn{6}{c}{\textsc{\bigLM}} \\
    \midrule
    
    \multirow{4}{*}{\textsc{2013}} & 2010 & 33.3 & 32.8 & 24.6 & 16.8 \\
    & 2012 & 21.4 & 39.5 & 30.0 & 14.1 \\
    & 2014 & 19.9 & 35.0 & 35.1 & 14.9 \\
    & 2016 & 19.9 & 21.2 & 21.1 & 25.7 \\
    \hline
    \multirow{4}{*}{\textsc{2016}} & 2010 & 31.9 & 33.3 & 27.1 & 17.8 \\
    & 2012 & 21.4 & 39.0 & 30.1 & 15.3 \\
    & 2014 & 20.2 & 35.0 & 34.5 & 17.2 \\
    & 2016 & 19.6 & 20.8 & 20.0 & 26.1 \\
    \hline
    \multirow{4}{*}{\textsc{2019}} & 2010 & 31.8 & 31.6 & 24.8 & 16.7 \\
    & 2012 & 21.4 & 39.1 & 29.3 & 13.6 \\
    & 2014 & 18.6 & 33.8 & 34.0 & 15.7 \\
    & 2016 & 19.5 & 20.1 & 21.4 & 26.2 \\
    \hline
    \multirow{4}{*}{\textsc{2022}} & 2010 & 30.7 & 30.8 & 24.4 & 17.2 \\
    & 2012 & 21.6 & 38.2 & 30.1 & 14.3 \\
    & 2014 & 19.5 & 35.5 & 35.0 & 14.7 \\
    & 2016 & 19.1 & 20.4 & 19.9 & 25.2 \\
    
    \midrule
    \multicolumn{6}{c}{\textsc{\smalLM}} \\
    \midrule
    
    \multirow{5}{*}{\textsc{2013}} & 2010 & 22.7 & 25.0 & 20.1 & 13.5 \\
    & 2012 & 14.0 & 24.5 & 19.5 & 9.9 \\
    & 2014 & 13.1 & 21.8 & 21.3 & 9.6 \\
    & 2016 & 14.1 & 17.8 & 17.5 & 18.4 \\
    \hline
    \multirow{5}{*}{\textsc{2016}} & 2010 & 22.1 & 25.5 & 20.7 & 14.0 \\
    & 2012 & 14.0 & 23.8 & 19.7 & 9.6 \\
    & 2014 & 13.5 & 22.8 & 21.5 & 10.0 \\
    & 2016 & 14.1 & 19.5 & 19.1 & 18.5 \\
    \hline
    \multirow{5}{*}{\textsc{2019}} & 2010 & 23.5 & 26.4 & 21.4 & 14.3 \\
    & 2012 & 14.5 & 25.4 & 20.6 & 10.1 \\
    & 2014 & 14.0 & 23.6 & 22.5 & 10.5 \\
    & 2016 & 15.1 & 20.1 & 19.2 & 18.5 \\
    \hline
    \multirow{5}{*}{\textsc{2022}} & 2010 & 23.4 & 26.2 & 21.1 & 14.1 \\
    & 2012 & 13.9 & 24.4 & 19.4 & 9.5 \\
    & 2014 & 13.6 & 23.2 & 21.7 & 9.7 \\
    & 2016 & 14.3 & 19.3 & 18.3 & 18.2 \\

    \bottomrule
    \end{tabular}
    \label{tab:time_newsum}
    \end{minipage}
\end{table}
\begingroup
\setlength{\tabcolsep}{4pt}
\begin{table*}[ht]
    \centering
    \small
    % \begin{tabular}{>{\raggedright}p{7cm} crll}
    \caption{
    Full results on the \textbf{TwiERC} temporal task splits from \citet{luu2021time}. This task evaluates Twitter Named Entity Classification with Accuracy.}
    \label{tab:time_twierc}
    \begin{tabular}[b]{ll | cccccc|cccccc}
    \toprule
    \textsc{Pretrain} & \textsc{Finetune} & \multicolumn{12}{c}{\textsc{Eval Time}} \\
    \textsc{Time} & \textsc{Time} & \textsc{2014} & \textsc{2015} & \textsc{2016} & \textsc{2017} & \textsc{2018} & \textsc{2019} & \textsc{2014} & \textsc{2015} & \textsc{2016} & \textsc{2017} & \textsc{2018} & \textsc{2019} \\

    \midrule
    & & \multicolumn{6}{c}{\textsc{\bigLM}} & \multicolumn{6}{c}{\textsc{\smalLM}} \\
    \midrule
    
    \multirow{3}{*}{\textsc{2013}} & 2014-2015 & 98.0 & 97.7 & 94.6 & 88.0 & 93.1 & 83.4 & 86.1 & 85.5 & 85.7 & 83.2 & 80.5 & 81.9 \\
    & 2016-2017 & 98.2 & 96.6 & 94.4 & 91.6 & 94.0 & 88.2 & 86.1 & 84.0 & 84.7 & 83.9 & 84.0 & 83.7 \\
    & 2018-2019 & 97.4 & 97.6 & 94.0 & 91.5 & 95.4 & 87.9 & 82.9 & 85.2 & 84.2 & 81.2 & 84.6 & 85.0 \\
    \hline
    \multirow{3}{*}{\textsc{2016}} & 2014-2015 & 98.4 & 98.3 & 95.1 & 87.5 & 92.5 & 82.7 & 86.2 & 85.7 & 86.2 & 82.7 & 81.5 & 81.7 \\
    & 2016-2017 & 97.8 & 97.5 & 94.6 & 91.9 & 93.3 & 86.7 & 86.7 & 84.1 & 86.0 & 85.1 & 83.2 & 83.4 \\
    & 2018-2019 & 96.7 & 98.0 & 94.1 & 91.3 & 95.7 & 87.6 & 82.7 & 84.6 & 85.5 & 81.5 & 85.5 & 85.0 \\
    \hline
    
    \multirow{3}{*}{\textsc{2019}} & 2014-2015 & 98.3 & 97.7 & 94.4 & 88.4 & 93.7 & 82.1 & 85.6 & 85.4 & 85.3 & 83.1 & 82.2 & 83.2 \\
    & 2016-2017 & 97.7 & 97.5 & 93.5 & 89.6 & 94.3 & 88.6 & 85.7 & 83.8 & 83.8 & 85.4 & 83.5 & 84.8 \\
    & 2018-2019 & 96.4 & 97.9 & 93.5 & 90.3 & 95.9 & 88.1 & 82.4 & 83.9 & 84.7 & 83.5 & 85.6 & 86.0 \\
    \hline
    \multirow{3}{*}{\textsc{2022}} & 2014-2015 & 98.4 & 98.1 & 95.1 & 88.1 & 94.1 & 84.6 & 84.4 & 84.8 & 85.6 & 83.0 & 82.0 & 81.7 \\
    & 2016-2017 & 97.9 & 97.2 & 93.8 & 89.4 & 94.6 & 88.3 & 83.2 & 83.1 & 84.5 & 83.1 & 82.2 & 83.6 \\
    & 2018-2019 & 96.5 & 97.6 & 93.9 & 90.7 & 96.3 & 87.9 & 80.5 & 83.1 & 83.2 & 82.6 & 84.0 & 85.7 \\

    \bottomrule
    \end{tabular}
    
\vspace{8em}

    \centering
    \small
    \caption{
    Full results on the \textbf{AIC} temporal task splits from \citep{luu2021time}. This task evaluates the classification of science articles from Semantic Scholar into those published at ICML or AAAI, measured with Accuracy.}
    \begin{tabular}{ll | cccccc|cccccc}
    \toprule
    \textsc{Pretrain} & \textsc{Finetune} & \multicolumn{12}{c}{\textsc{Eval Time}} \\
    \textsc{Time} & \textsc{Time} & \textsc{2014} & \textsc{2015} & \textsc{2016} & \textsc{2017} & \textsc{2018} & \textsc{2019} & \textsc{2014} & \textsc{2015} & \textsc{2016} & \textsc{2017} & \textsc{2018} & \textsc{2019} \\

    \midrule
    & & \multicolumn{6}{c}{\textsc{\bigLM}} & \multicolumn{6}{c}{\textsc{\smalLM}} \\
    \midrule
    
    \multirow{3}{*}{\textsc{2013}} & 2014-2015 & 98.7 & 97.5 & 95.6 & 89.0 & 94.0 & 86.0 & 74.5 & 75.3 & 80.4 & 74.0 & 71.9 & 69.5 \\
    & 2016-2017 & 98.2 & 98.0 & 95.0 & 93.1 & 95.2 & 90.2 & 74.3 & 74.0 & 77.0 & 75.4 & 74.7 & 70.9 \\
    & 2018-2019 & 97.7 & 98.5 & 94.4 & 91.8 & 94.0 & 89.9 & 68.1 & 70.2 & 76.2 & 71.2 & 75.4 & 75.0 \\
    \hline
    \multirow{3}{*}{\textsc{2016}} & 2014-2015 & 98.5 & 98.4 & 95.6 & 92.0 & 94.3 & 86.3 & 74.9 & 75.9 & 81.7 & 74.4 & 71.0 & 70.7 \\
    & 2016-2017 & 98.0 & 98.1 & 95.4 & 94.0 & 95.1 & 89.7 & 74.1 & 72.9 & 78.9 & 74.0 & 74.1 & 70.0 \\
    & 2018-2019 & 97.6 & 98.2 & 94.6 & 93.4 & 95.8 & 89.4 & 69.5 & 70.3 & 76.7 & 72.1 & 75.3 & 75.3 \\
    \hline
    \multirow{3}{*}{\textsc{2019}} & 2014-2015 & 98.2 & 98.5 & 95.0 & 93.6 & 94.8 & 88.0 & 74.9 & 75.9 & 79.4 & 76.8 & 70.3 & 69.7 \\
    & 2016-2017 & 97.9 & 98.8 & 94.0 & 94.0 & 96.4 & 91.4 & 73.9 & 74.5 & 78.4 & 75.0 & 74.9 & 69.7 \\
    & 2018-2019 & 97.3 & 98.9 & 92.7 & 92.5 & 96.7 & 92.0 & 67.8 & 69.8 & 77.5 & 73.9 & 75.4 & 76.2 \\
    \hline
    \multirow{3}{*}{\textsc{2022}} & 2014-2015 & 98.2 & 97.4 & 95.3 & 90.6 & 94.2 & 87.3 & 72.8 & 78.6 & 78.3 & 72.6 & 70.7 & 69.5 \\
    & 2016-2017 & 97.5 & 98.9 & 94.7 & 91.7 & 95.8 & 90.9 & 71.9 & 73.4 & 77.6 & 74.4 & 72.6 & 69.0 \\
    & 2018-2019 & 97.0 & 98.8 & 93.1 & 91.9 & 95.1 & 88.7 & 66.8 & 71.6 & 74.6 & 73.9 & 74.7 & 72.7 \\
    \bottomrule
    \end{tabular}
    \label{tab:time_aic}

\end{table*}
\endgroup

\begingroup
\setlength{\tabcolsep}{4pt}
\begin{table*}[ht]
    \centering
    \small
    % \begin{tabular}{>{\raggedright}p{7cm} crll}
    \caption{
    Full results on the \textbf{PoliAff} temporal task splits from \citet{luu2021time}. This task evaluates classification of political affiliation from tweets, measured in Accuracy.}
    \begin{tabular}{ll | cccccccccc}
    \toprule
    \textsc{Pretrain} & \textsc{Finetune} & \multicolumn{10}{c}{\textsc{Eval Time}} \\
    \textsc{Time} & \textsc{Time} & \textsc{2012} & \textsc{2013} & \textsc{2014} & \textsc{2015} & \textsc{2016} & \textsc{2017} & \textsc{2018} & \textsc{2019} & \textsc{2020} & \textsc{2021} \\
    % \multirow{2}{*}{\textsc{Pretrain\\Time}} & \multirow{2}{*}{\textsc{Finetune\\Time}} & \multicolumn{10}{c}{\textsc{Eval Time}} \\
    % & & \textsc{2012} & \textsc{2013} & \textsc{2014} & \textsc{2015} & \textsc{2016} & \textsc{2017} & \textsc{2018} & \textsc{2019} & \textsc{2020} & \textsc{2021} \\
    \midrule
    \multicolumn{12}{c}{\textsc{\bigLM}} \\
    \midrule

    \multirow{5}{*}{\textsc{2013}} & 2009-2013 & 100.0 & 100.0 & 95.5 & 73.5 & 65.4 & 56.5 & 51.1 & 70.1 & 74.2 & 67.2 \\
    & 2014-2015 & 95.1 & 97.9 & 100.0 & 97.4 & 81.9 & 65.2 & 56.7 & 72.6 & 73.0 & 66.4 \\
    & 2016-2017 & 88.2 & 92.8 & 95.0 & 87.2 & 92.3 & 92.8 & 80.4 & 82.6 & 79.0 & 69.3 \\
    & 2018-2019 & 76.8 & 86.0 & 88.4 & 58.9 & 66.2 & 87.0 & 91.3 & 94.5 & 90.2 & 79.9 \\
    & 2020-2021 & 53.6 & 68.4 & 77.0 & 39.2 & 48.1 & 72.0 & 77.9 & 90.2 & 94.7 & 91.9 \\
    \hline
    \multirow{5}{*}{\textsc{2016}} & 2009-2013 & 100.0 & 100.0 & 94.9 & 72.9 & 67.8 & 55.8 & 53.7 & 70.3 & 73.7 & 67.5 \\
    & 2014-2015 & 94.9 & 98.2 & 100.0 & 97.3 & 82.5 & 68.4 & 58.7 & 73.0 & 73.4 & 67.9 \\
    & 2016-2017 & 85.0 & 92.6 & 94.6 & 87.8 & 91.8 & 93.1 & 80.7 & 83.0 & 79.9 & 69.2 \\
    & 2018-2019 & 73.1 & 85.2 & 87.9 & 58.3 & 68.5 & 88.1 & 91.3 & 94.4 & 90.4 & 81.5 \\
    & 2020-2021 & 49.0 & 64.3 & 75.5 & 38.0 & 50.8 & 73.4 & 78.6 & 90.5 & 94.6 & 93.7 \\
    \hline
    \multirow{5}{*}{\textsc{2019}} & 2009-2013 & 100.0 & 100.0 & 95.5 & 73.3 & 68.0 & 57.9 & 55.2 & 71.8 & 74.3 & 68.4 \\
    & 2014-2015 & 93.8 & 97.4 & 100.0 & 97.7 & 82.5 & 69.7 & 59.1 & 74.6 & 73.9 & 67.9 \\
    & 2016-2017 & 85.0 & 92.7 & 94.6 & 87.1 & 92.0 & 93.1 & 82.0 & 83.4 & 80.4 & 68.3 \\
    & 2018-2019 & 73.8 & 84.8 & 87.6 & 58.4 & 68.9 & 86.7 & 91.9 & 94.8 & 90.3 & 81.4 \\
    & 2020-2021 & 48.4 & 64.2 & 75.6 & 35.7 & 48.6 & 71.7 & 78.6 & 90.7 & 95.0 & 93.7 \\
    \hline
    \multirow{5}{*}{\textsc{2022}} & 2009-2013 & 100.0 & 100.0 & 94.9 & 72.6 & 67.5 & 55.6 & 52.3 & 72.0 & 76.7 & 69.0 \\
    & 2014-2015 & 94.4 & 97.9 & 100.0 & 97.9 & 81.0 & 68.0 & 56.1 & 74.3 & 73.9 & 68.8 \\
    & 2016-2017 & 84.1 & 91.5 & 93.2 & 85.2 & 90.9 & 92.2 & 78.7 & 80.9 & 79.7 & 69.6 \\
    & 2018-2019 & 71.1 & 83.9 & 86.0 & 58.9 & 66.6 & 85.8 & 90.3 & 94.5 & 91.1 & 83.0 \\
    & 2020-2021 & 47.2 & 62.4 & 73.0 & 39.6 & 50.8 & 73.6 & 77.5 & 90.6 & 94.9 & 93.8 \\

    \midrule
    \multicolumn{12}{c}{\textsc{\smalLM}} \\
    \midrule
    
    \multirow{5}{*}{\textsc{2013}} & 2009-2013 & 89.1 & 87.5 & 80.2 & 48.5 & 42.3 & 38.9 & 42.4 & 57.0 & 62.9 & 56.4 \\
    & 2014-2015 & 77.8 & 88.5 & 89.5 & 64.7 & 50.4 & 46.3 & 42.0 & 60.3 & 63.3 & 55.7 \\
    & 2016-2017 & 40.9 & 43.4 & 58.2 & 36.1 & 40.0 & 54.7 & 47.4 & 61.2 & 61.2 & 54.4 \\
    & 2018-2019 & 41.2 & 39.3 & 44.0 & 21.7 & 23.0 & 42.3 & 49.8 & 63.1 & 67.2 & 56.9 \\
    & 2020-2021 & 40.8 & 37.9 & 42.6 & 20.5 & 22.5 & 37.2 & 45.4 & 64.6 & 71.9 & 65.6 \\
    \hline
    \multirow{5}{*}{\textsc{2016}} & 2009-2013 & 89.9 & 89.2 & 80.5 & 51.7 & 45.7 & 39.9 & 42.6 & 57.7 & 62.6 & 55.4 \\
    & 2014-2015 & 78.2 & 87.8 & 87.4 & 63.9 & 49.6 & 45.6 & 41.8 & 59.7 & 61.9 & 54.3 \\
    & 2016-2017 & 51.3 & 49.3 & 57.9 & 37.4 & 38.1 & 51.1 & 46.3 & 60.2 & 60.2 & 53.6 \\
    & 2018-2019 & 49.8 & 43.1 & 46.5 & 24.4 & 26.8 & 42.6 & 48.3 & 62.9 & 66.3 & 56.2 \\
    & 2020-2021 & 51.7 & 43.0 & 42.5 & 22.7 & 24.8 & 36.3 & 40.8 & 61.5 & 70.1 & 63.3 \\
    \hline
    \multirow{5}{*}{\textsc{2019}} & 2009-2013 & 89.2 & 87.0 & 77.9 & 48.5 & 39.8 & 38.7 & 41.7 & 57.8 & 64.6 & 55.6 \\
    & 2014-2015 & 73.3 & 87.7 & 87.9 & 63.8 & 48.7 & 42.8 & 39.5 & 57.4 & 61.8 & 53.8 \\
    & 2016-2017 & 34.8 & 45.7 & 55.6 & 36.6 & 36.2 & 50.1 & 44.5 & 59.8 & 60.4 & 53.1 \\
    & 2018-2019 & 32.6 & 36.4 & 43.6 & 21.6 & 21.7 & 41.2 & 48.7 & 62.8 & 66.6 & 55.7 \\
    & 2020-2021 & 34.8 & 37.6 & 43.7 & 21.3 & 21.3 & 36.0 & 42.4 & 62.7 & 70.9 & 62.0 \\
    \hline
    \multirow{5}{*}{\textsc{2022}} & 2009-2013 & 90.3 & 88.8 & 79.0 & 47.9 & 41.0 & 37.6 & 40.9 & 57.9 & 64.7 & 56.6 \\
    & 2014-2015 & 76.9 & 89.7 & 90.3 & 67.2 & 54.6 & 45.2 & 41.0 & 60.5 & 63.4 & 56.5 \\
    & 2016-2017 & 41.5 & 48.8 & 56.9 & 37.0 & 38.6 & 53.7 & 47.7 & 62.0 & 60.7 & 53.2 \\
    & 2018-2019 & 33.0 & 34.3 & 39.2 & 19.9 & 20.5 & 43.2 & 50.9 & 65.5 & 68.8 & 56.4 \\
    & 2020-2021 & 39.5 & 37.0 & 38.5 & 19.4 & 19.6 & 33.6 & 41.8 & 65.2 & 72.8 & 66.1 \\

    \bottomrule
    \end{tabular}
    \label{tab:time_poliaff}
\end{table*}
\endgroup

\clearpage
\vspace{-3mm}
\subsection{Toxicity \& Quality Filtering Results}

\begin{figure}[ht]
    \centering
    \includegraphics[width=.4\textwidth]{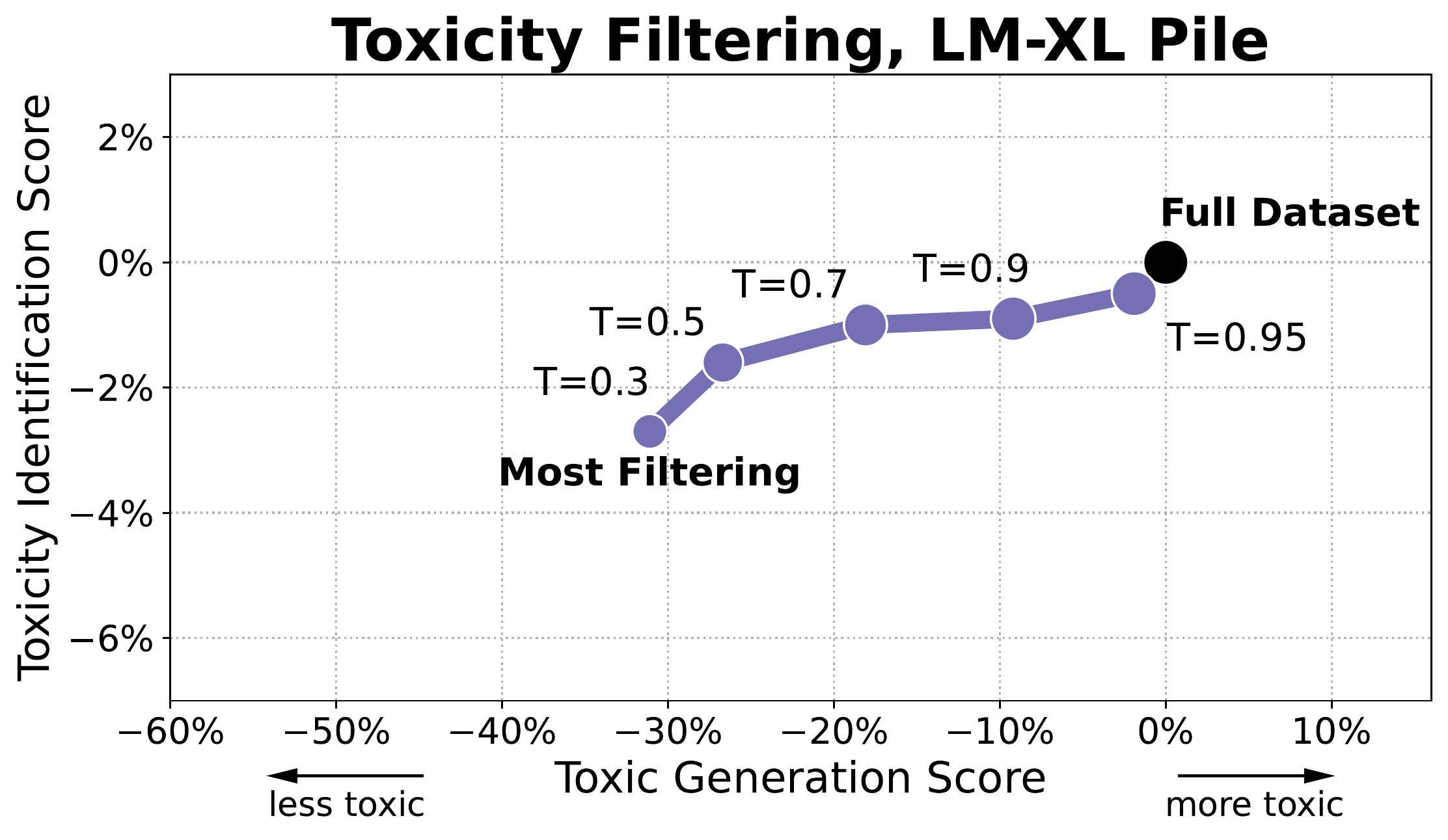}
    \caption{
    \textbf{Toxicity filtering the Pile decreases the ability of \bigLM to identify toxicity and to generate toxic text, just as with toxicity filtering C4.}
    }
    \label{fig:tox_filter_pile}
\end{figure}

We also provide full results for our experiments with toxicity and quality filters, presented in \cref{sec:quality-toxicity}.
The evaluation results of the models with \emph{toxicity} filters applied to their data are visualized in \cref{fig:tox_qual_filter_c4} (left) and \cref{fig:tox_filter_pile}, with full details in \cref{tab:tox-filter-results}.
The evaluation results of the models with \emph{quality} filters applied to their data are visualized in \cref{fig:tox_qual_filter_c4} (right) and detailed in \cref{tab:qual-filter-results}.

\begingroup
\setlength{\tabcolsep}{4pt}
\begin{table*}[ht]
    \centering
    \small
    
    \caption{
    \textbf{Toxicity filtering the pre-training dataset decreases the ability of \bigLM to identify toxicity and to generate toxic text.} These results are visualized in Figures~\ref{fig:tox_qual_filter_c4} and \ref{fig:tox_filter_pile}.}
    \label{tab:tox-filter-results}
    
    % \begin{tabular}{>{\raggedright}p{7cm} crll}
    \begin{tabular}{l | r | cccc | r || ccc | r}
    \toprule
    \textsc{Filter} & \textsc{\% Data} & \multicolumn{5}{c}{\textsc{Toxicity Identification ($\uparrow$)}} & \multicolumn{4}{c}{\textsc{Toxicity Generation ($\downarrow$)}} \\
    & & SBF & Toxigen & DH R3 & DH R4 & Score & RTP-T & RPT-NT & RepBias & Score \\
    \midrule
    \multicolumn{11}{c}{\textsc{The Pile}} \\
    \midrule
    
    \textsc{Full Dataset} & 100.0 & 90.7 & 90.8 & 88.7 & 84.1 & 0.0 & 88.9 & 44.4 & 4.6$\pm$0.7 & 0.0 \\
    \textsc{T=0.95} & 99.1 & 90.6 & 90.9 & 87.8 & 83.5 & \cellcolor{color3!5} -0.5 & 85.6 & 43.9 & 4.6$\pm$0.8 & \cellcolor{color3!20} -1.9 \\
    \textsc{T=0.9} & 97.4 & 90.2 & 90.8 & 86.4 & 83.7 & \cellcolor{color3!10} -0.9 & 80.4 & 41.9 & 4.0$\pm$0.6 & \cellcolor{color3!28} -9.2 \\
    \textsc{T=0.7} & 90.8 & 89.9 & 90.9 & 87.4 & 82.7 & \cellcolor{color3!10} -1.0 & 83.3 & 39.9 & 2.9$\pm$0.5 & \cellcolor{color3!40} -18.1 \\
    \textsc{T=0.5} & 80.7 & 89.4 & 90.4 & 86.0 & 82.8 & \cellcolor{color3!20} -1.6 & 83.3 & 35 & 2.2$\pm$0.4 & \cellcolor{color3!52} -26.7 \\
    \textsc{T=0.3} & 60.1 & 88.4 & 89.9 & 85.3 & 81.3 & \cellcolor{color3!25} -2.7 & 78.5 & 31.4 & 2.2$\pm$0.5 & \cellcolor{color3!62} -31.1 \\
    \textsc{Ngrams} & 70.7 & 89.7 & 90.4 & 86.3 & 82.4 & \cellcolor{color3!20} -1.6 & 76.1 & 33.6 & 2.5$\pm$0.6 & \cellcolor{color3!56} -28.0 \\
    
    \midrule
    \multicolumn{11}{c}{\textsc{C4}} \\
    \midrule

    \textsc{Inverse T=0.06} & 92.2 & 93.2 & 91.4 & 90.0 & 85.7 & \cellcolor{forestgreen!14} 1.4 & 87.8 & 49.6 & 4.8$\pm$0.8 & \cellcolor{forestgreen!50} 15.6 \\
    \textsc{Full Dataset} & 100.0 & 91.2 & 91.1 & 89.0 & 84.2 & 0.0 & 84.6 & 41.8 & 3.9$\pm$0.7 & 0.0 \\
    \textsc{T=0.95} & 97.7 & 90.7 & 91.3 & 87.7 & 83.4 & \cellcolor{color3!7}-0.7 & 84.3 & 41.9 & 3.9$\pm$0.7 & 0.0 \\
    \textsc{T=0.9} & 94.9 & 90.4 & 90.6 & 87.5 & 83.9 & \cellcolor{color3!10}-0.9 & 81.1 & 40.3 & 3.1$\pm$0.6 & \cellcolor{color3!28}-9.0 \\
    \textsc{T=0.7} & 85.8 & 90.5 & 90.5 & 86.1 & 82.8 & \cellcolor{color3!16}-1.6 & 71.3 & 34.8 & 2.4$\pm$0.5 & \cellcolor{color3!47}-23.8 \\
    \textsc{T=0.5} & 75.8 & 89.8 & 90.5 & 86.9 & 81.9 & \cellcolor{color3!18}-1.8 & 65.2 & 30.0 & 1.8$\pm$0.4 & \cellcolor{color3!70}-35.0 \\
    \textsc{T=0.3} & 60.8 & 89.4 & 90.2 & 82.1 & 75.6 & \cellcolor{color3!25}-5.2 & 55.0 & 19.8 & 1.2$\pm$0.3 & \cellcolor{color3!80}-52.1 \\
    \textsc{Ngrams} & 78.6 & 89.8 & 90.7 & 87.0 & 81.8 & \cellcolor{color3!18}-1.8 & 74.7 & 31.8 & 2.3$\pm$0.5 & \cellcolor{color3!51}-25.6 \\

    \bottomrule
    \end{tabular}

\end{table*}
\endgroup

\begingroup
\setlength{\tabcolsep}{4pt}
\begin{table*}[ht]
    \centering
    \small
    \caption{
    \textbf{Quality filtering the pre-training dataset decreases the ability of \bigLM to identify toxicity but surprisingly increases toxicity generation.}
    These results are visualized in \cref{fig:tox_qual_filter_c4}.}
    \label{tab:qual-filter-results}
    
    \begin{tabular}{l | r | cccc | r || ccc | r}
    \toprule
    \textsc{Filter} & \textsc{\% Data} & \multicolumn{5}{c}{\textsc{Toxicity Identification ($\uparrow$)}} & \multicolumn{4}{c}{\textsc{Toxicity Generation ($\downarrow$)}} \\
    & & SBF & Toxigen & DH R3 & DH R4 & Score & RTP-T & RPT-NT & RepBias & Score \\
    \midrule
    \multicolumn{11}{c}{\textsc{C4}} \\
    \midrule
  
    \textsc{Inverse T=0.5} & 73.3 & 91.8 & 90.1 & 86.8 & 82.9 & \cellcolor{color3!9}{-0.9} & 86.3 & 44.3 & 4.1$\pm$0.6 & \cellcolor{forestgreen!41}{+9.7} \\
    \textsc{Full Dataset} & 100.0 & 93.1 & 91.0 & 87.4 & 83.5 & 0.0 & 84.1 & 41.8 & 3.4$\pm$0.6 & 0.0 \\
    \textsc{T=0.975} & 90.6 & 93.1 & 91.3 & 87.8 & 82.7 & \cellcolor{color3!3}{-0.1} & 85.4 & 46.0 & 3.8$\pm$0.7 & \cellcolor{forestgreen!16}{+7.3} \\
    \textsc{T=0.95} & 83.9 & 93.2 & 91.3 & 89.4 & 85.0 & \cellcolor{forestgreen!11}{+1.1} & 86.3 & 44.0 & 4.2$\pm$0.6 & \cellcolor{forestgreen!47}{+10.4} \\
    \textsc{T=0.9} & 73.3 & 93.3 & 91.2 & 88.6 & 85.9 & \cellcolor{forestgreen!12}{+1.2} & 85.2 & 44.8 & 4.3$\pm$0.7 & \cellcolor{forestgreen!60}{+11.1} \\
    \textsc{T=0.7} & 45.6 & 93.3 & 91.4 & 89.9 & 86.6 & \cellcolor{forestgreen!18}{+1.8} & 86.5 & 44.7 & 4.0$\pm$0.8 & \cellcolor{forestgreen!40}{+9.6} \\

    \bottomrule
    \end{tabular}
\end{table*}
\endgroup

\end{document}